\documentclass[10pt]{article} 

\usepackage{wrapfig}
\usepackage{graphicx}
\usepackage{amsfonts}
\usepackage{amssymb}
\usepackage{amsbsy}
\usepackage{amsmath}
\usepackage{multirow}
\usepackage{array}
\usepackage{color}
\usepackage{soul}
\usepackage{cite}
\usepackage[normalem]{ulem} 
\usepackage{times}
\usepackage[all]{xy}
\usepackage{lscape}
\usepackage{flushend}
\usepackage{url}
\usepackage{float}
\usepackage{rotating}

\def\mH{\mathcal{H}}

\def\Real{\mathbb{R}}

\def\K{\ensuremath{\mathbf{K}}}
\def\L{\ensuremath{\mathbf{L}}}

\def\U{\ensuremath{\mathbf{U}}}
\def\V{\ensuremath{\mathbf{V}}}
\def\W{\ensuremath{\mathbf{W}}}
\def\X{\ensuremath{\mathbf{X}}}

\def\Z{\ensuremath{\mathbf{Z}}}

\def\aa{\boldsymbol{\alpha}}
\def\pphhii{\boldsymbol{\phi}}
\def\PHI{\boldsymbol{\Phi}}
\def\LL{\boldsymbol{\Lambda}}

\def\u{\ensuremath{\mathbf{u}}} 
\def\v{\ensuremath{\mathbf{v}}}

\def\u{\ensuremath{\mathbf{u}}}
\def\x{\ensuremath{\mathbf{x}}}

\newcommand{\red}[1]{\textcolor[rgb]{0.8,0,0}{#1}}           
\newcommand{\green}[1]{\textcolor[rgb]{0,0.8,0}{#1}}           
\newcommand{\blue}[1]{\textcolor[rgb]{0,0,1}{#1}}           
\newcommand{\cyan}[1]{\textcolor[rgb]{0,0.8,0.8}{#1}}           
\newcommand{\devis}[2]{\textcolor[rgb]{0,0.8,0.2}{#2}}     

\newtheorem{theorem}{\bf Theorem}

\title{Kernel Manifold Alignment}

\author{
Devis Tuia \\
Department of Geography, University of Zurich, Switzerland \\
\texttt{devis.tuia@geo.uzh.ch} \\
Gustau~Camps-Valls \\
Image Processing Laboratory, Universitat de Val\`encia, Spain \\
\texttt{gustau.camps@uv.es} \\
}

\begin{document}

\maketitle

\begin{abstract}
We introduce a kernel method for manifold alignment (KEMA) and domain adaptation that can 
match an arbitrary number of data sources without needing  
corresponding pairs, just few labeled examples in all domains. KEMA has interesting properties: 
1) it generalizes other manifold alignment methods, 
2) it can align manifolds of very different complexities, performing a sort of manifold unfolding plus alignment, 
3) it can define a domain-specific metric to cope with multimodal specificities, 
4) it can align data spaces of different dimensionality, 
5) it is robust to strong nonlinear feature deformations, and
6) it is closed-form invertible which allows transfer across-domains and data synthesis. 
To authors' knowledge this is the first method in addressing all these important issues at once. 
We also present a reduced-rank version for computational efficiency and 
discuss the generalization performance of KEMA under Rademacher principles of stability. 
KEMA exhibits very good performance over competing methods in  
synthetic examples, visual object recognition and 
recognition of facial expressions tasks.
\end{abstract}

\section{Introduction}

Domain adaptation constitutes a field of high interest in pattern analysis and machine learning. Classification algorithms developed with data from one domain cannot be directly used in another related domain, and hence adaptation of either the classifier or the data representation become strictly imperative
~\cite{Quinonero09}. 
In this paper, we focus on the latter pathway, which has been referred to as \emph{feature representation transfer}~\cite{Pan09}, \emph{feature transformation learning}~\cite{Pat14} or \emph{manifold alignment}~\cite{Wan11}. Roughly speaking, aligning data manifolds reduces to finding projections to a common latent space where all datasets show similar statistical characteristics. 
Depending on the availability of labels in the different domains, three families of adaptation problems have been considered in the literature. 

\emph{Unsupervised adaptation:} First attempts of unsupervised domain adaptation are found in multiview analysis~\cite{Jac12}, and more precisely in canonical correlation analysis (CCA) and kernel CCA (KCCA)~\cite{Lai00}. Despite their good performance in general, they still require points in different sources to be corresponding pairs, which is often hard to meet in real applications.
Alternative methods seek for a set of projectors that minimize a measure of discrepancy between the source and target data distributions, such as the  Maximum Mean Discrepancy (MMD)~\cite{Pan11} or the recent geodesic distance between distributions~\cite{Bak14}. 
However, to compare distributions, the data are supposed to be represented by the same features in all domains.
The idea of exploiting geodesic distances along manifolds was also considered in~\cite{gopalan11}, where a finite set of intermediate transformed data distributions are sampled along the geodesic flow (SGF) between the linear subspaces. The intermediate features are then used to train the classifier. The idea was extended in~\cite{gong12}, where a Geodesic Flow Kernel (GFK) was constructed by considering the infinity of transformed subspaces along the geodesic path. However, both SGF and GFK assume input data space of the same dimensionality. 

\emph{Semi-supervised adaptation with labels in the source domain only:} 
Some of the abovementioned methods can incorporate the information of labeled samples in the source domain: 
For example, SGF~\cite{gopalan11} and GFK~\cite{gong12} become semi-supervised if the eigenvectors of the source domain are found with a discriminative feature extractor such as partial least squares (PLS). 
Another family of methods, collectively known as Optimal Transport (OT) techniques, uses labeled samples in the source domain to maximize coherence in the transportation plan of masses between source and target domains~\cite{Cou14}. 

\emph{Supervised adaptation with labels in all domains:} SGF and GFK can be also defined for the case in which all the domains are labeled. 
Alternative approaches try to align target and source features while simultaneously moving labeled examples to the correct side of the decision hyperplane (MMDT)~\cite{Hof13}. 
A last family of supervised methods is known as {\em manifold alignment}, and aims at 
concurrently matching the corresponding instances while preserving the topology of each input domain, generally using a graph Laplacian~\cite{Ham05,Wan11}. While appealing, these methods still require specifying a small amount of cross-domain sample correspondences. The problem was addressed in~\cite{Wang11} by relaxing the constraint of paired correspondences with the constraint of having the same class labels in all domains.  
The semi-supervised manifold alignment (SSMA) method proposed in~\cite{Wang11} projects data from different domains to a latent space where samples belonging to the same class become closer, those of different classes are pushed far apart, and the geometry of each domain is preserved. The method performs well in general and can deal with multiple domains of different dimensionality. However, SSMA cannot cope with strong nonlinear deformations and high-dimensional data problems. 


This paper introduces a generalization of SSMA through kernelization. The proposed Kernel Manifold Alignment (KEMA) has appealing properties: 
(1) it reduces to SSMA when using a linear kernel, which allows us to deal with high-dimensional data efficiently in the dual form ($Q$-mode analysis): by this property, KEMA can cope with input space of very large dimension, e.g. those extracted by Fisher vectors or deep features;  
(2) it goes beyond data rotations so it can align manifolds of very different structures, performing a sort of manifold unfolding simultaneous to the alignment;  
(3) it can also define a domain-specific metric by the use of different kernel functions in the different domains; 
(4) as SSMA, KEMA can align data spaces of different dimensionality; 
(5) it is robust to strong (nonlinear) deformations of the manifolds to be aligned, as the kernel compensates for problems in graph estimation and numerical problems; and 
(6) mapping inversion (and hence data synthesis) can be performed in closed-form without the need of pre-images, which permits measuring the quality of the alignment in meaningful physical units.

The remainder of the paper is organized as follows. Section 2 briefly reviews the main properties of the SSMA algorithm. Section 3 introduces the KEMA formulation and analyzes its theoretical and practical properties. Section 4 presents the experimental evaluation of the algorithm. We compare KEMA to SSMA and related (linear and kernel) methods in toy examples and real visual object and face recognition problems. 
We conclude with some remarks in Section 5.

\section{Semi-supervised Manifold Alignment}

Let us consider $D$ domains ${\mathcal X}_i$ representing similar classification problems. The corresponding data matrices, $\X_i\in \Real^{{d_i}\times n_i}$, $i=1,\ldots, D$, contain $n_i$ examples (labeled, $l_i$, and unlabeled, $u_i$, with $n_i = l_i + u_i$) of dimension ${d_i}$, and $n=\sum_{i=1}^D n_i$. The SSMA method~\cite{Wang11} maps all the data to a latent space ${\mathcal F}$ such that samples belonging to the same class become closer, those of different classes are pushed far apart, and the geometry of the data manifolds is preserved. Therefore, three entities have to be considered, leading to three $n \times n$ matrices: 1) a similarity matrix $\W_s$ that has components $W_s^{ij}=1$ if $\x_i$ and $\x_j$ belong to the same class, and 0 otherwise (including unlabeled); 2) a dissimilarity matrix $\W_d$, which has entries $W_d^{ij}=1$ if $\x_i$ and $\x_j$ belong to different classes, and $0$ otherwise (including unlabeled); and 3) a similarity matrix that represents the topology of a given domain, $\W$, e.g. a radial basis function (RBF) kernel or a $k$ nearest neighbors graph computed for each domain separatedly and joined in a block-diagonal matrix. 
The three different entities lead to three different graph Laplacians: $\L_s$, $\L_d$, and $\L$, respectively. Then, the SSMA embedding must minimize a joint cost function essentially given by the eigenvectors corresponding to the smallest non-zero eigenvalues of the following generalized eigenvalue problem:
$$\Z (\L + \mu \L_s)\Z^\top \V = \lambda \Z \L_d \Z^\top \V,$$
where $\Z$ is a block diagonal matrix containing the data matrices $\X_i$ and $\V$ contains in the columns the eigenvectors organized in rows for the particular domain, $\V =[\v_1, \v_2, \ldots, \v_D]^\top$. The method allows to extract a maximum of $N_f=\sum_{i=1}^D {d_i}$ features that serve for projecting the data to the common latent domain as follows:
$$P_{\mathcal F}(\X_i) = \v_i^\top \X_i.$$

Advantageously, SSMA can easily project data between domains $j$ and $i$: first mapping the data in ${\mathcal X}_j$ to the latent domain ${\mathcal F}$, and from there inverting back to the target domain ${\mathcal X}_i$ as follows:
$$P_i(\X_j) = (\v_j\v_i^\dag)^\top \X_j,$$
where $^\dagger$ represents the pseudo-inverse of the eigenvectors of the target domain. 
Therefore, the method can be used for domain adaptation but also for data synthesis. 

\section{Kernel manifold alignment}

In order to kernelize the previous method 
one needs to first map the data to a Hilbert space, apply the representer's theorem and replace the dot products therein with reproducing kernel functions.  Let us first map the $D$ different datasets to $D$ possibly different Hilbert spaces $\mH_i$ of dimension $H_i$, $\pphhii_i(\cdot):\x\mapsto\pphhii_i(\x)\in \mH_i$, $i=1,\ldots,D$. Now, by replacing all the samples with their mapped feature vectors, the problem becomes:
$$\PHI (\L + \mu \L_s)\PHI^\top \U = \lambda \PHI \L_d \PHI^\top \U,$$
where $\PHI$ is a block diagonal matrix containing the data matrices $\PHI_i=[\pphhii_i(\x_1), \ldots, \pphhii_i(\x_{n_i})]^\top$ and $\U$ contains the eigenvectors organized in rows for the particular domain defined in Hilbert space $\mH_i$, $\U =[\u_1, \u_2, \ldots, \u_H]^\top$ where $H=\sum_i^D H_i$. This operation is possible thanks to the use of the direct sum of Hilbert spaces, a well-known property of Functional Analysis Theory~\cite{ReedSimon81}. 
Note that the eigenvectors $\u_i$ are of possibly infinite dimension and cannot be explicitly computed. Instead, we resort to the definition of $D$ corresponding Riesz representation theorems~\cite{RieNag55} 
so the eigenvectors can be expressed as a linear combination of mapped samples~\cite{Yan07}, $\u_i=\PHI_i\aa_i$, and in matrix notation $\U = \PHI\LL$. This leads to the problem:
\begin{equation}\label{primalKEMA}
\PHI (\L + \mu \L_s)\PHI^\top \PHI\LL = \lambda \PHI \L_d \PHI^\top \PHI\LL.
\end{equation}
Now, by premultiplying both sides by $\PHI^\top$ 
and replacing the dot products with the corresponding kernel matrices, $\K_i = \PHI_i^\top\PHI_i$, we obtain the final solution:
$$\K (\L + \mu \L_s)\K\LL = \lambda \K \L_d \K \LL,$$
where $\K$ is a block diagonal matrix containing the kernel matrices $\K_i$. Now the eigenproblem becomes of size $n\times n$ instead of $d\times d$, and we can extract a maximum of $N_f=n$ features. 

\paragraph{Kernel generalization:}  
When a linear kernel is used for all the domains, $\K_i=\X_i^\top \X_i$, KEMA reduces to SSMA:
$$P_{\mathcal F}(\X_i) = \aa_i^\top \X_i^\top \X_i = (\X_i\aa_i)^\top \X_i = \v_i^\top \X_i.$$
This dual formulation is advantageous when dealing with very high dimensional datasets, $d_i\gg n_i$ for which the SSMA problem is not well-conditioned. Operating in $Q$-mode endorses the method with numerical stability and computational efficiency in current high-dimensional problems, e.g. when using Fisher vectors or deep features. 

\paragraph{Projections to kernel latent space: } 
Projection to the latent space requires first mapping the data $\X_i$ to its corresponding Hilbert space $\mH_i$, thus leading to the mapped data $\PHI_i$, and then applying the projection vector $\u_i$ defined therein:
\begin{equation}
P_{\mathcal F}(\X_i) = \u_i^\top\PHI_i = \aa_i^\top\PHI_i^\top\PHI_i = \aa_i^\top \K_i.
\label{eq:proj}
\end{equation}

\paragraph{Invertibility has a closed-form solution: }
In order to map data from ${\mathcal X}_j$ to ${\mathcal X}_i$ with KEMA we would need to estimate $D-1$ inverse mappings, which would make KEMA unstable and useless to measure accuracy in meaningful physical units of the input space. 
In general, using kernel functions hampers the invertibility of the transformation unless pre-imaging is used, for which some efficient yet inexact solutions exist~\cite{BakWesSch03,KwoTsa04}.
Here we propose a simple closed-form solution to the mapping inversion: to use a linear kernel for the latent-to-target transformation 
$\K_i=\X_i^\top \X_i$, and $\K_j$ for $j\neq i$ with any desired form. 
Then, projection of data $\X_j$ to the target domain $i$ becomes:
\begin{equation}
	P_i(\X_j)=(\u_i^\dag)^\top\aa_j^\top \K_j=(\aa_j (\X_i\aa_i)^\dag)^\top \K_j,
	\label{eq:inv}
\end{equation}
where for the target domain we used $\u_i =  \PHI_i\aa_i =  \X_i\aa_i$. 
We should note that the solution is not unique since $D$ different inverse solutions can be obtained depending on the selected target domain. 

\subsection{Reduced rank approximation}\label{sec:rekema}

KEMA complexity scales quadratically with $n$ in terms of memory, and cubically with respect to the computation time. Feature extraction for new data requires the evaluation of $n$ kernel functions {\em per} pattern, becoming computationally expensive for large $n$. To alleviate this problem, we propose a reduced-rank approximation of the span. The so-called Reduced-Rank Kernel Manifold Alignment (REKEMA) formulation imposes {reduced}-rank solutions for the projection vectors, $\W = \PHI_r\LL$, where $\PHI_r$ is a subset of the training data containing $r$ samples ($r \ll n$) and $\LL$ is the new argument for the maximization problem. 
Plugging $\W$ into Eq.~\eqref{primalKEMA}, and replacing the dot products with the corresponding kernels, $\K_{rn} = \PHI_r^\top\PHI$, we obtain the final solution:
$$\K_{rn} (\L + \mu \L_s)\K_{nr}\LL = \lambda \K_{rn} \L_d \K_{nr} \LL,$$
where $\K_{rn}$ is a block diagonal matrix containing the kernel matrices $\K_i$ comparing a reduced set of $r$ representative vectors and {\em all} training data points, $n$. REKEMA reports clear benefits for obtaining the projection vectors (eigenproblem becomes of size $r\times r$ instead of $n\times n$), compacting the solution (now $N_f=r\ll n$ features), and in storage requirements (quadratic with $r$).

\subsection{Stability of KEMA}

The use of KEMA in practice raises the question of the amount of data needed to provide an accurate empirical estimate and how the quality of the solution differs depending on the datasets. Such results have been previously derived for KPCA~\cite{jst05} and KPLS~\cite{Dhanjal09} and here we adapt them to our setting. The following properties are based on the concentration of sums of eigenvalues of the generalized KEMA eigenproblem solved using a finite number of samples, where new points are projected into the $m$-dimensional space spanned by the $m$ eigenvectors corresponding to the largest $m$ eigenvalues. Following the notation in \cite{jst05}, we refer to the projection onto a subspace $U$ of the eigenvectors of our eigenproblem as $P_{U}(\phi({\bf x}))$. We represent the projection onto the orthogonal complement of $U$ by $P_{{\bf U}^{\perp}}(\phi({\bf x}))$. The norm of the orthogonal projection is also referred to as the residual since it corresponds to the distance between the points and their projections.

\begin{theorem}[Th. 1 and 2 in \cite{jst05}]
If we perform KEMA in the feature space defined by ${\bf K}^* = ({\bf K}({\bf L}+\mu{\bf L}_s){\bf K})^{-1}{\bf K}{\bf L}_d{\bf K}$, then with probability greater than $1-\delta$ over $n$ random samples $S$, for all $1\leq m\leq n$, if we project data on the space $\hat{U}_m$, the expected squared residual is bounded by
$$
\displaystyle
\sum_{j=m+1}^{n} \lambda_j \leq \mathbb{E} \left[  {\| P_{{\hat{U}_m}^\perp}\|}^2  \right]  
\leq   \min_{1\leq l \leq m}\left[    \frac{1}{n} \sum_{j=l+1}^{n} \hat{\lambda}_j(S) +\frac{1+\sqrt{l}}{\sqrt{n}} \sqrt{  \frac{2}{n}\sum_{i=1}^{n} K_{ii}^{*2}  } \right] 
 +    R^2\sqrt{\frac{18}{n}\ln\left(  \frac{2n}{\delta}  \right)}  
$$
and
$$
\displaystyle
\sum_{j=1}^{m} \lambda_j \leq \mathbb{E} \left[  {\| P_{\hat{U}_m}\|}^2  \right] 
\leq   \max_{1\leq l \leq m}\left[    \frac{1}{n} \sum_{j=1}^{l} \hat{\lambda}_j(S) -\frac{1+\sqrt{l}}{\sqrt{n}} \sqrt{  \frac{2}{n}\sum_{i=1}^{n} K_{ii}^{*2}  } \right] 
 -    R^2\sqrt{\frac{19}{n}\ln\left(  \frac{2(n+1)}{\delta}  \right)},
$$
where the support of the distribution is in a ball of radius $R$ in the feature space and $\lambda_i$ are $\hat{\lambda}_i$ are the process and empirical eigenvalues, respectively.
\end{theorem}
The lower bound confirms that a good representation of the data can be achieved by using the first $m$ eigenvectors if the empirical eigenvalues quickly decrease before $\sqrt{l/n}$ becomes large, while the upper bound suggests that a good approximation 
is achievable for values of $m$ where $\sqrt{m/n}$ is small.
These results can be used as a benchmark to test different approaches or to select among possible candidate kernels. Also, note that depending on how much non-diagonal is $\K^\ast$ (i.e. how large are the manifold mis-alignments), the KEMA bounds may be tighter than those of KPCA. With an appropriate estimation of the manifold structures via the graph Laplacians and tuning of the kernel parameters, the performance of KEMA will be at least as fitted as that of KPCA.

\section{Experimental results}

We analyze the behavior of KEMA in a series of artificial datasets of controlled level of distortion and mis-alignment, and on real domain adaptation problems of visual object recognition from multi-source commercial databases, and recognition of multi-subject facial expressions.

\subsection{Toy examples with controlled distortions and manifold mis-alignments}

\begin{figure*}[h!]
\small
\begin{center}
\setlength{\tabcolsep}{0pt}
\begin{tabular}{|ccc|cc|cc|cc|cc|}
\hline
&&& \multicolumn{2}{c}{Input spaces} & \multicolumn{2}{c}{KEMA (linear kernel)} & \multicolumn{2}{c|}{KEMA (RBF kernel)}& \multicolumn{2}{c|}{Error rates [\%]}\\
\cline{3-9}
Exp. &&& Domains  & Classes & Domains  & Classes & Domains  & Classes & Domain $\sharp$1 &  Domain $\sharp$2 \\
\hline\hline
\rotatebox{90}{\hspace{0.2cm} Exp. \#1}& 
\rotatebox{90}{\hspace{0.3cm}$d_{\red{\bullet}} = 2$}& 
\rotatebox{90}{\hspace{0.3cm}$d_{\blue{\bullet}} = 2$}& 
\includegraphics[height = 1.55cm]{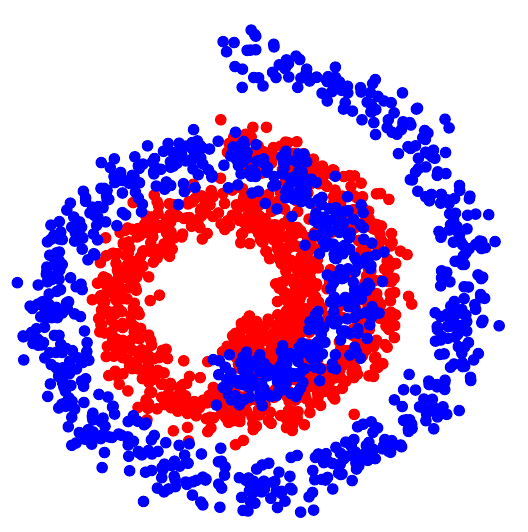}&
\includegraphics[height = 1.55cm]{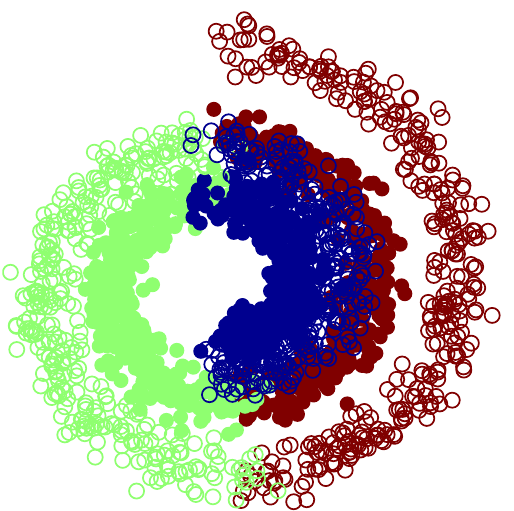}&
\includegraphics[height = 1.55cm]{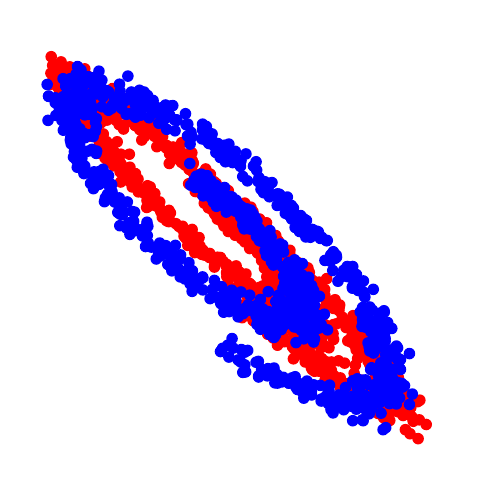}&
\includegraphics[height = 1.55cm]{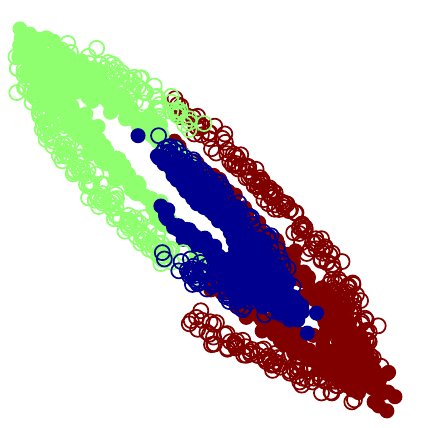}&
\includegraphics[height = 1.55cm]{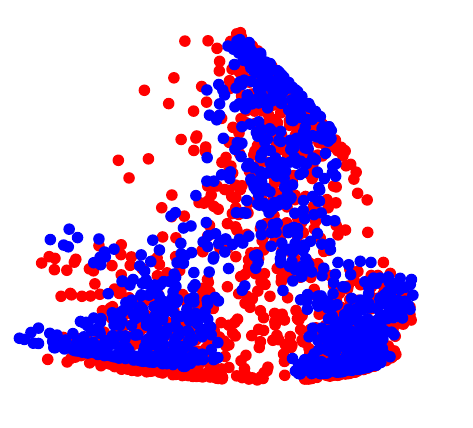}&
\includegraphics[height = 1.55cm]{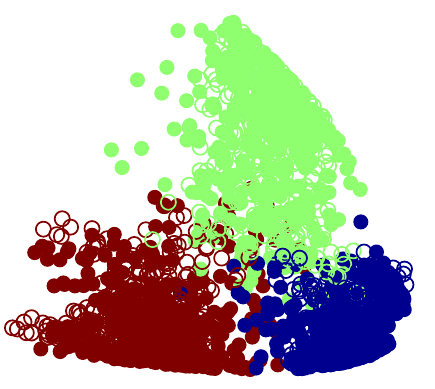}&
\includegraphics[height = 1.55cm]{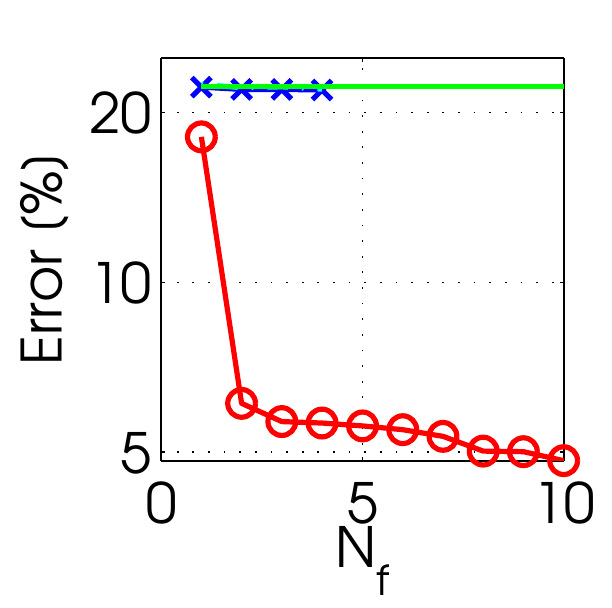}&
\includegraphics[height = 1.55cm]{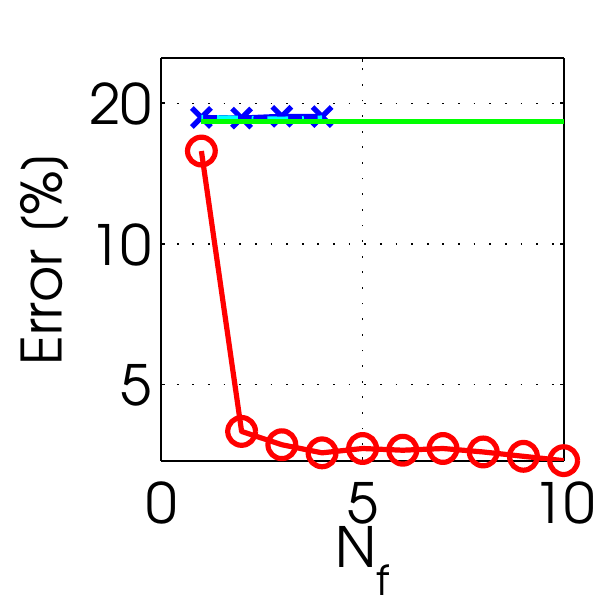}\\\hline


\rotatebox{90}{\hspace{0.2cm} Exp. \#2}& 
\rotatebox{90}{\hspace{0.3cm}$d_{\red{\bullet}} = 3$}& 
\rotatebox{90}{\hspace{0.3cm}$d_{\blue{\bullet}} = 2$}& 
\includegraphics[height = 1.55cm]{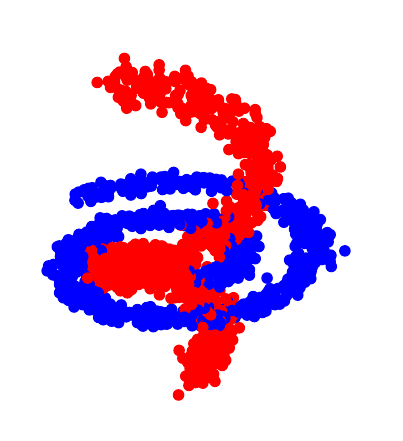}&
\includegraphics[height = 1.55cm]{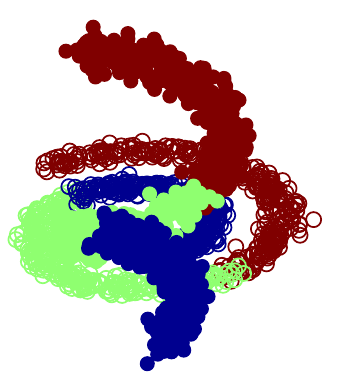}&
\includegraphics[height = 1.55cm]{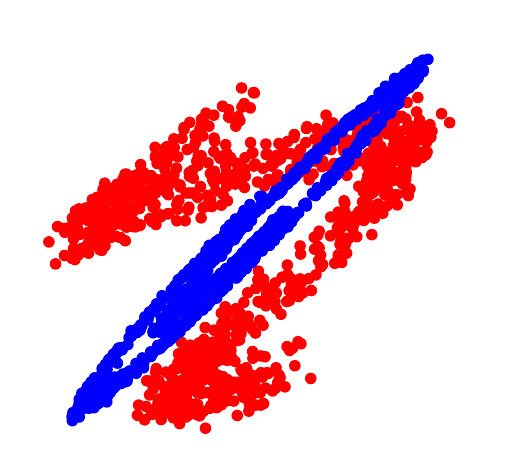}&
\includegraphics[height = 1.55cm]{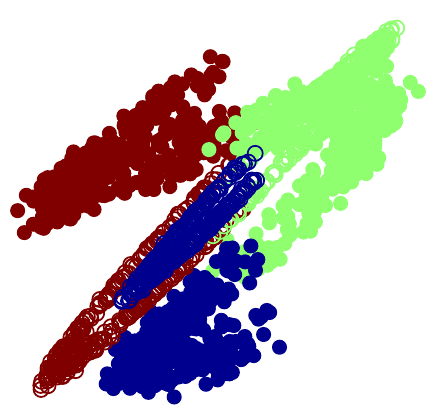}&
\includegraphics[height = 1.55cm]{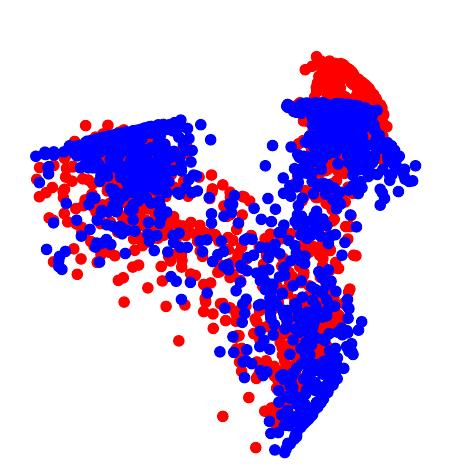}&
\includegraphics[height = 1.55cm]{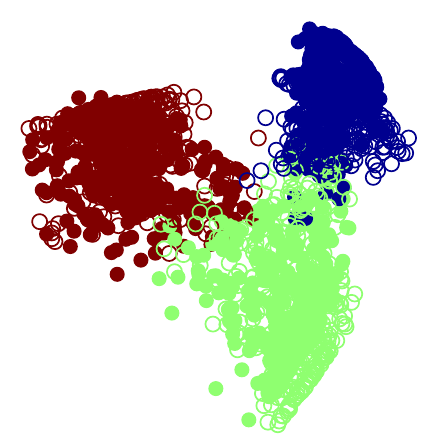}&
\includegraphics[height = 1.55cm]{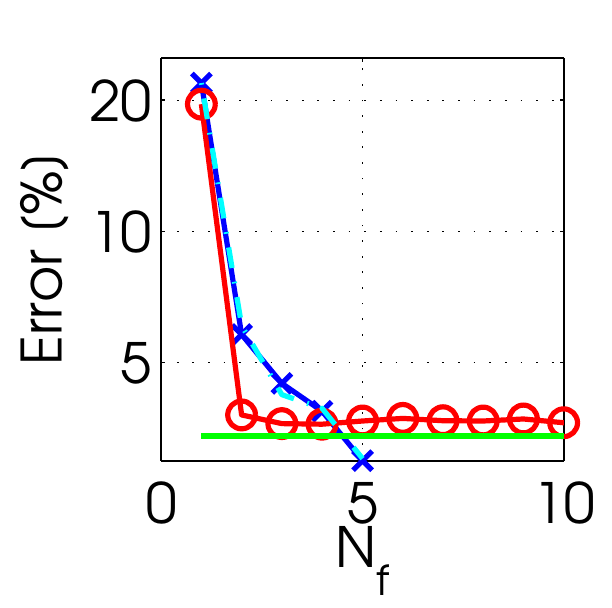}&
\includegraphics[height = 1.55cm]{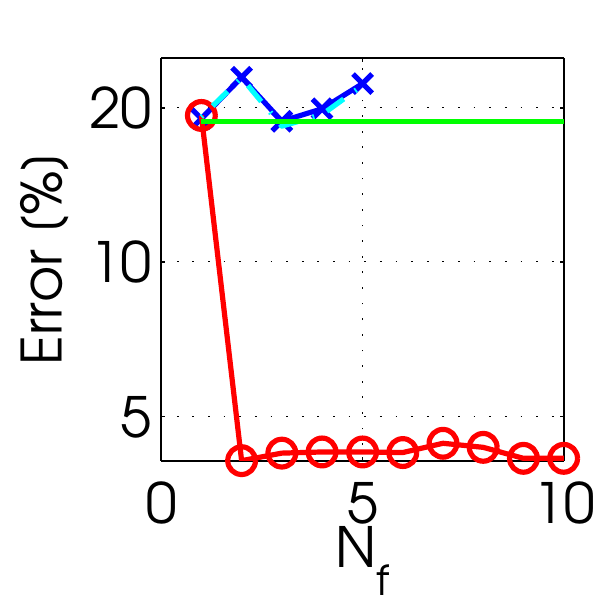}\\\hline

\rotatebox{90}{\hspace{0.2cm} Exp. \#3}& 
\rotatebox{90}{\hspace{0.3cm}$d_{\red{\bullet}} = 3$}& 
\rotatebox{90}{\hspace{0.3cm}$d_{\blue{\bullet}} = 3$}& 
\includegraphics[height = 1.55cm]{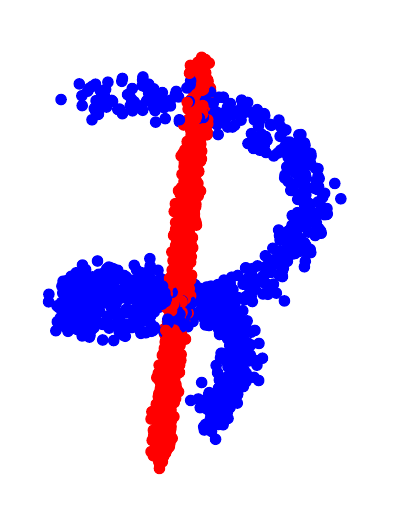}&
\includegraphics[height = 1.55cm]{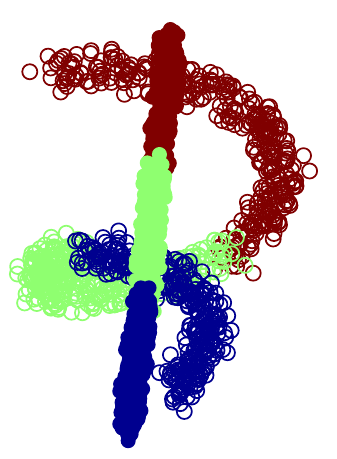}&
\includegraphics[height = 1.55cm]{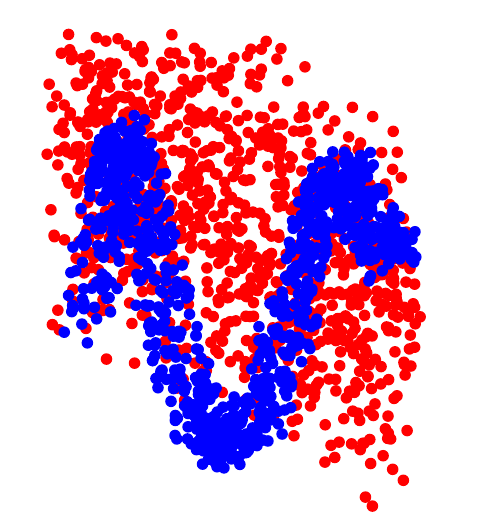}&
\includegraphics[height = 1.55cm]{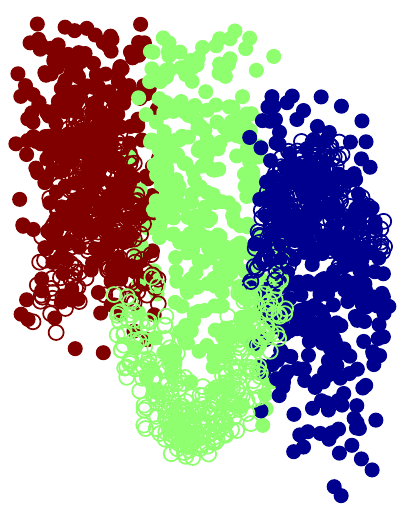}&
\includegraphics[height = 1.55cm]{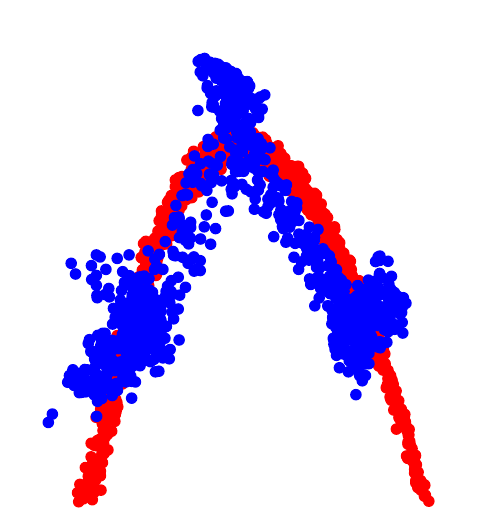}&
\includegraphics[height = 1.55cm]{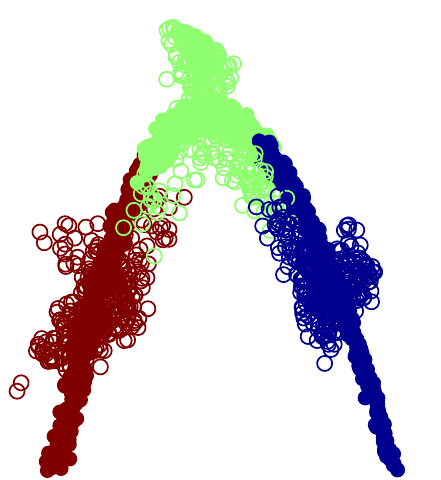}&
\includegraphics[height = 1.55cm]{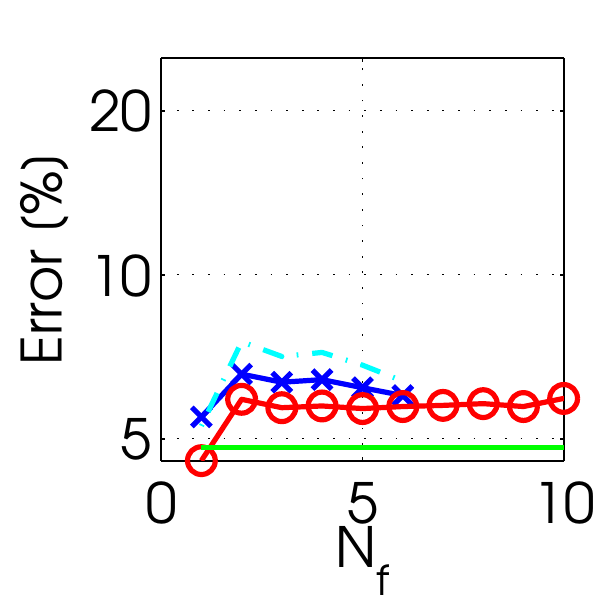}&
\includegraphics[height = 1.55cm]{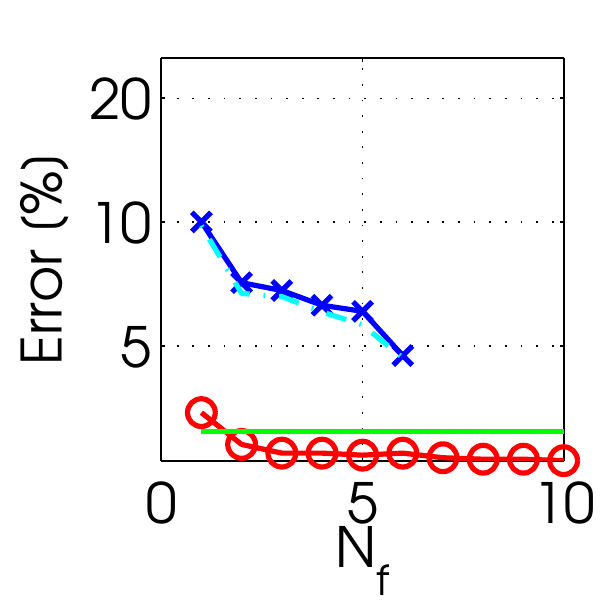}\\\hline

\rotatebox{90}{\hspace{0.2cm} Exp. \#4}& 
\rotatebox{90}{\hspace{0.3cm}$d_{\red{\bullet}} = 3$}& 
\rotatebox{90}{\hspace{0.3cm}$d_{\blue{\bullet}} = 3$}&  
\includegraphics[height = 1.55cm]{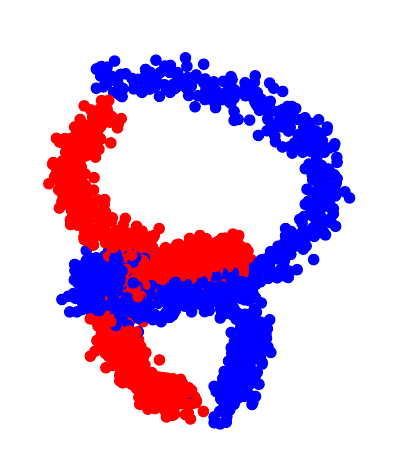}&
\includegraphics[height = 1.55cm]{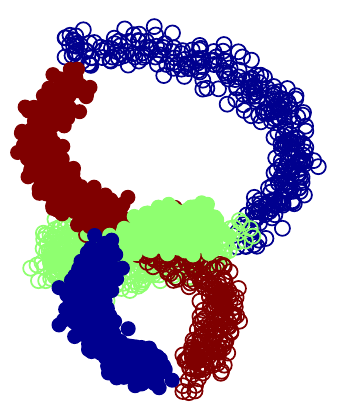}&
\includegraphics[height = 1.55cm]{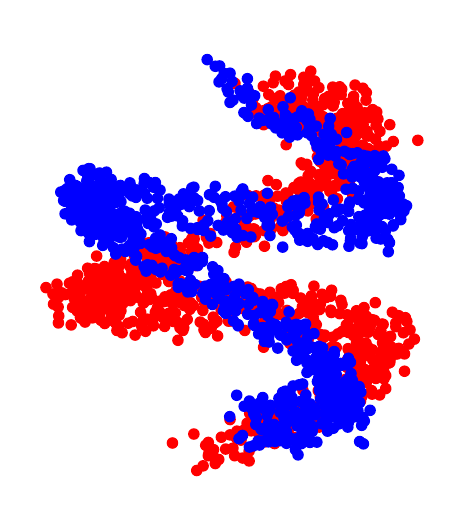}&
\includegraphics[height = 1.55cm]{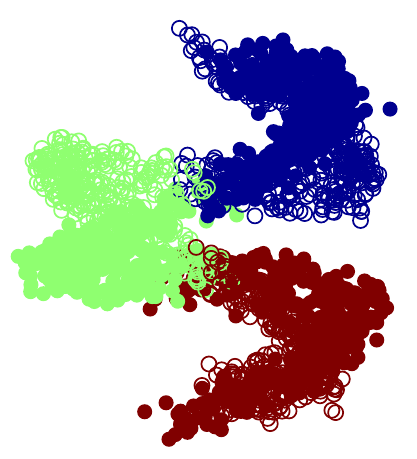}&
\includegraphics[height = 1.55cm]{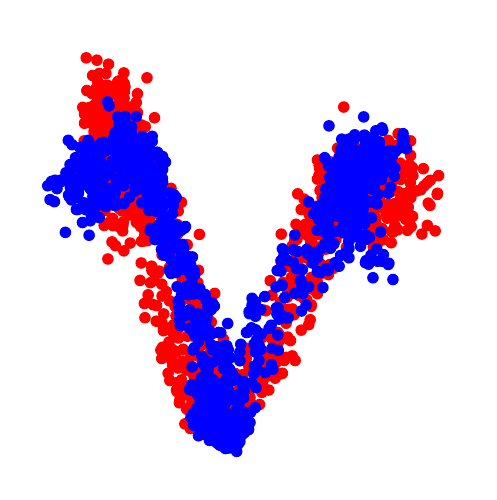}&
\includegraphics[height = 1.55cm]{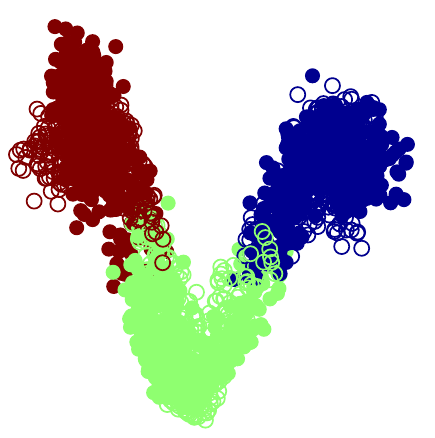}&
\includegraphics[height = 1.55cm]{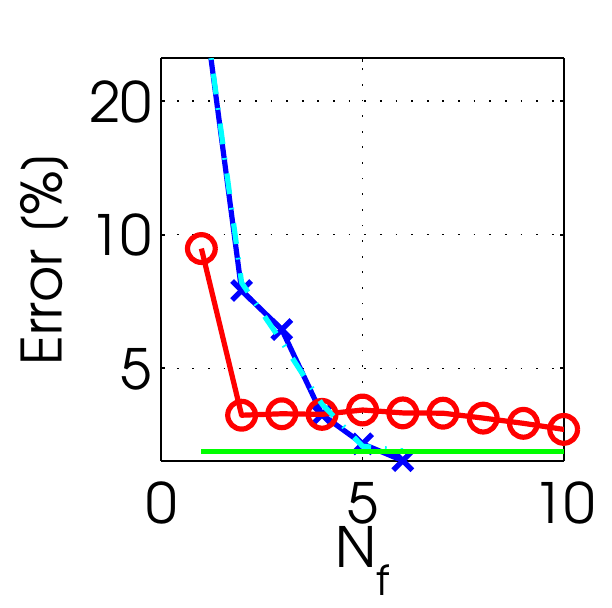}&
\includegraphics[height = 1.55cm]{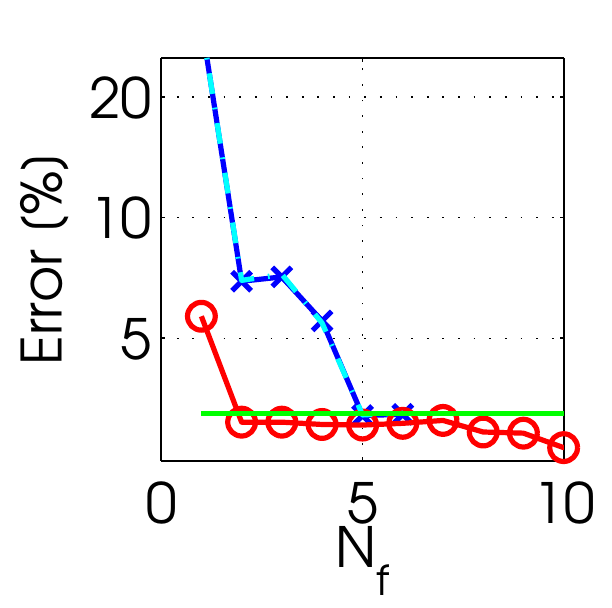}\\\hline

\hline
\end{tabular}
\caption{Illustration of linear and kernel manifold alignment on the toy experiments. Left to right: data in the original domains (X1 = $\red{\bullet}$, X2 = $\blue{\bullet}$) and {\em per} class ($\red{\bullet}$, $\green{\bullet}$ and $\blue{\bullet}$), data projected with the linear and the RBF kernels, and error rates as a function of the extracted features when predicting data for the first (left inset) or the second (right inset) domain (\blue{KEMA$_{\text{Lin}}$}, \red{KEMA$_{\text{RBF}}$}, \cyan{SSMA}, \green{Baseline}).}
\label{fig:toy}
\end{center}
\end{figure*}

\paragraph{Setup:} the first battery of experiments contains a series of toy examples composed of two domains with data matrices $\X_1$ and $\X_2$, which are spirals with three classes (see the two first columns of Fig.~\ref{fig:toy}). Then, a series of deformations are applied to the second domain: scaling, rotation, inversion of the order of the classes, the shape of the domain (spiral or line) or the data dimensionality. 
For each experiment, 20 labeled pixels {\em per} class were sampled in each domain, as well as 1000 unlabeled samples that were randomly selected. Classification performance was assessed on 1000 held-out samples from each domain.

\paragraph{Latent space and domain adaptation:} Figure~\ref{fig:toy} illustrates the projections obtained by KEMA when using a linear and an RBF kernel (lengthscale was set as the average distance between labeled samples) and the classification errors for the samples from the source domain (7$th$ column) and the target (8$th$ column). The linear KEMA (SSMA) can align effectively the domains in experiments \#1 and \#4, which are basically scalings and rotations of the data. However, it fails on experiments \#2 and \#3, where the manifolds have undergone stronger deformations. The use of a nonlinear kernel allows much more flexible solution, performing a sort of unfolding plus alignment in all experiments. In experiment \#1, even if the alignment is correct, the linear classifier trained on the projections of KEMA$_{\text{lin}}$ and SSMA cannot resolve the classification of the two domains, while KEMA$_{\text{RBF}}$ solution provides a latent space where both domains can be classified correctly. Experiment \#2 shows a different picture: the baseline error (green line) is much smaller in the source domain, since the dataset in 3D is linearly separable. Even if the classification of this first domain ($\red{\bullet}$) is correct for all methods, classification after SSMA/KEMA$_{\text{lin}}$ projection of the second domain ($\blue{\bullet}$) is poor, since their projection in the latent space does not unfold the blue spiral. KEMA$_{\text{RBF}}$ provides the best result. For experiment \#3, the same trend as in experiment \#2 is observed. Finally, experiment \#4  shows a very accurate baseline (both domains are linearly separable in the input spaces) and all methods provide accurate classification accuracies. Again, KEMA$_{\text{RBF}}$ provides the best match between the domains in the latent space.

\begin{figure*}[h!]
\small
\setlength{\tabcolsep}{5pt}
\begin{tabular}{p{1.3cm}|c|c|cccc|c}
\hline
Alignment&
No adapt. & 
SSMA  & 
\multicolumn{4}{c|}{REKEMA$_{\text{RBF}}$} & 
KEMA$_{\text{RBF}}$\\
\hline
$r/n \times 100$ & 100\%& 100\%& $1\%$  & $10\%$    & $25\%$    & $50\%$    & $100\%$  \\
SVM$_{\text{LIN}}$ & 72.33\% & 81.83\% & 50.83\%  & 97.17\% & 97.17\% & 97.67\% & 97.83\%\\
SVM$_{\text{RBF}}$    & 83.21\% & 95.12\% & 55.17\%& 98.12\% & 98.12\% & 98.87\% & 98.89\%\\
&
\includegraphics[height = 1.55cm]{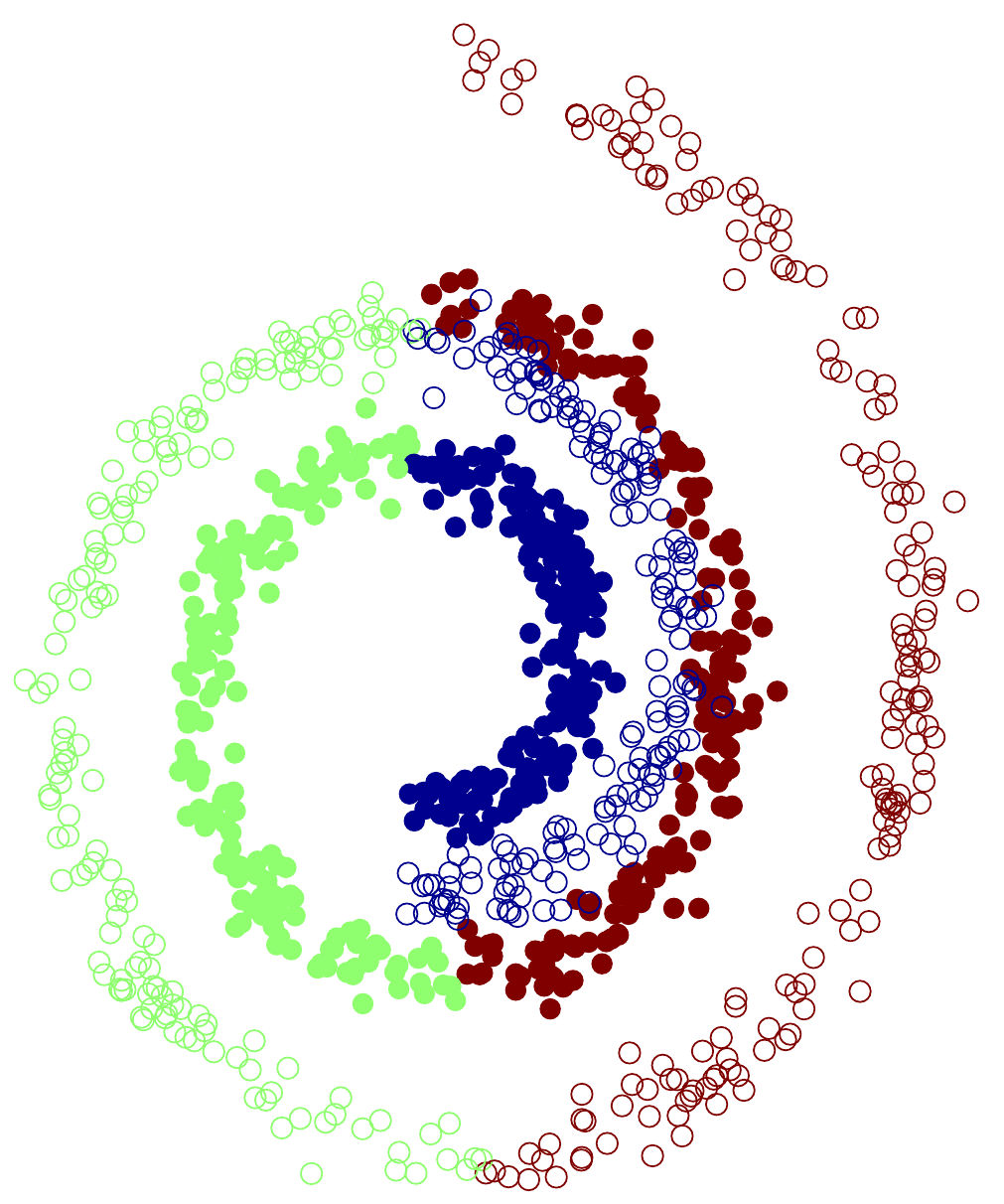}&
\includegraphics[height = 1.55cm]{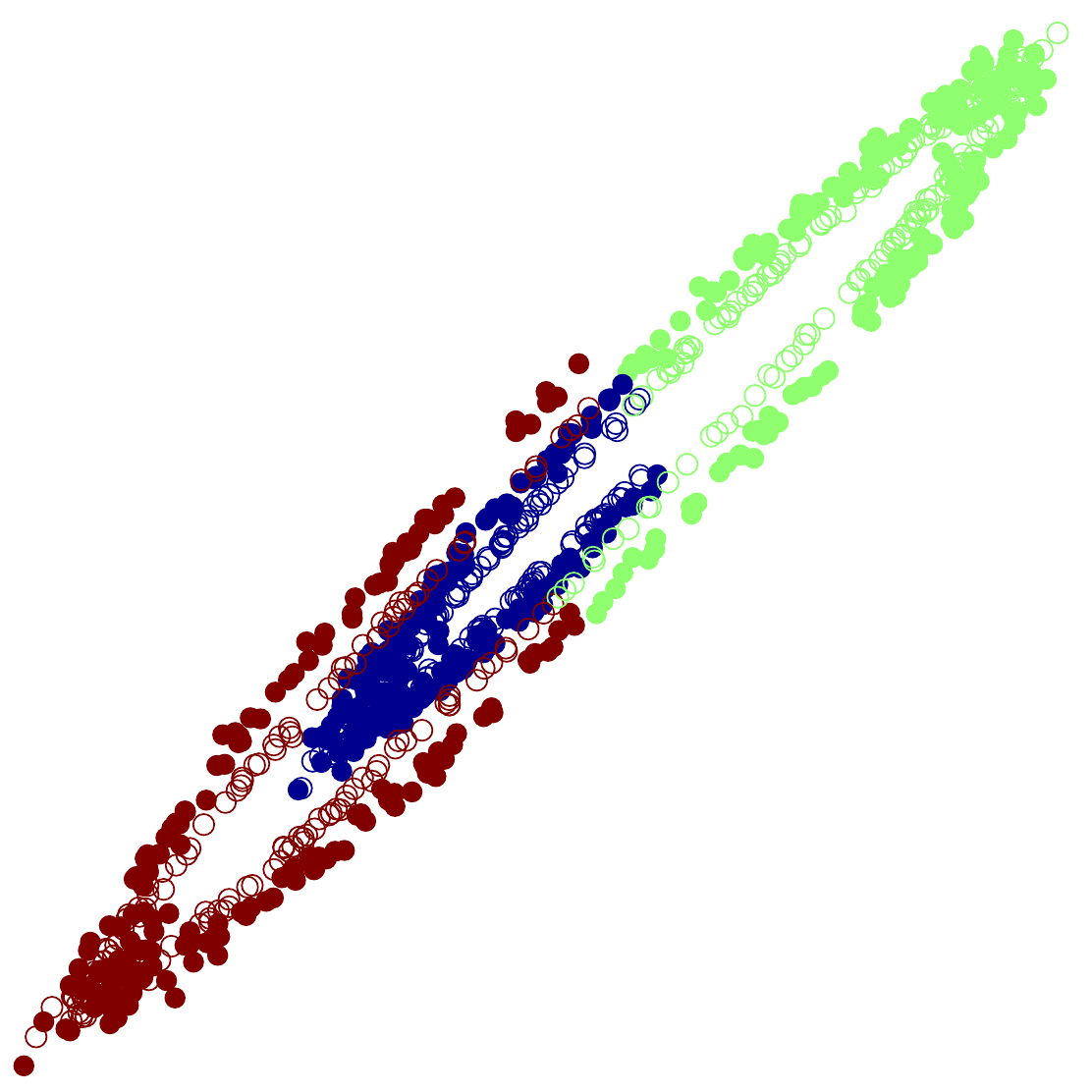}&
\includegraphics[height = 1.55cm]{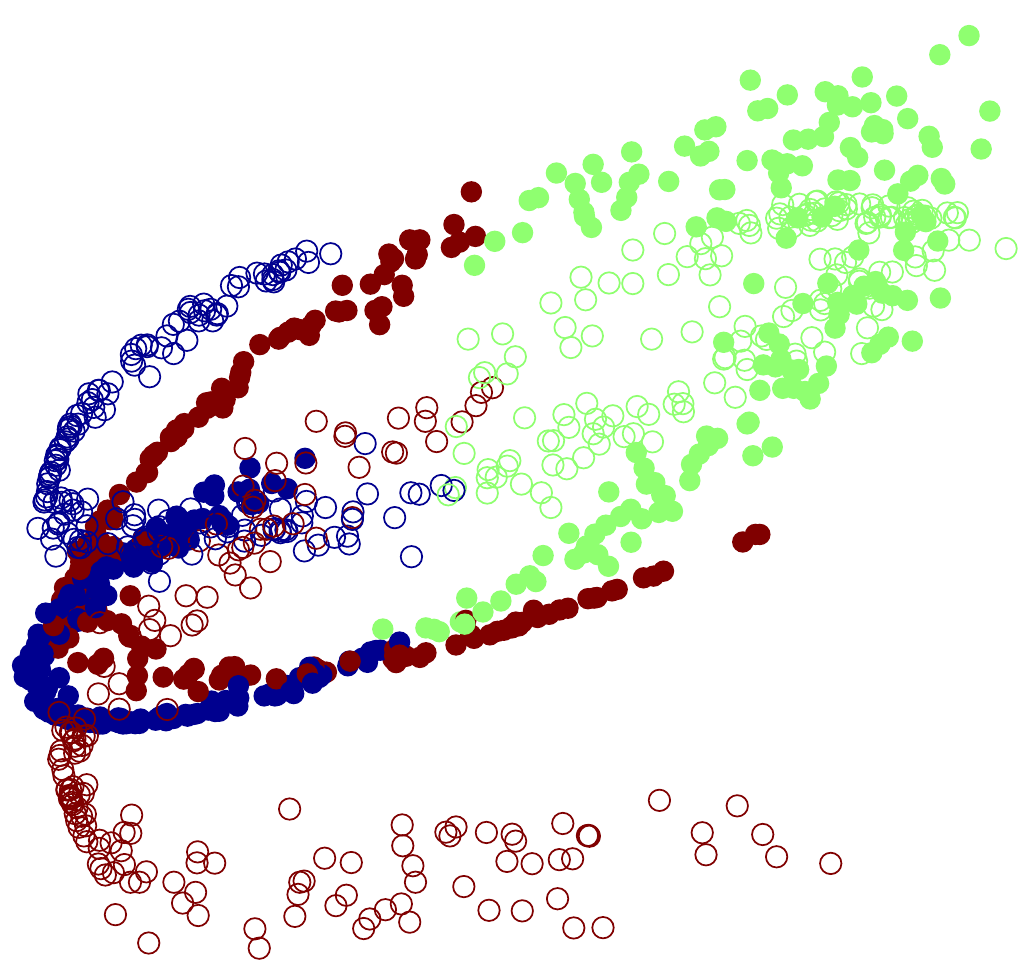}&
\includegraphics[height = 1.55cm]{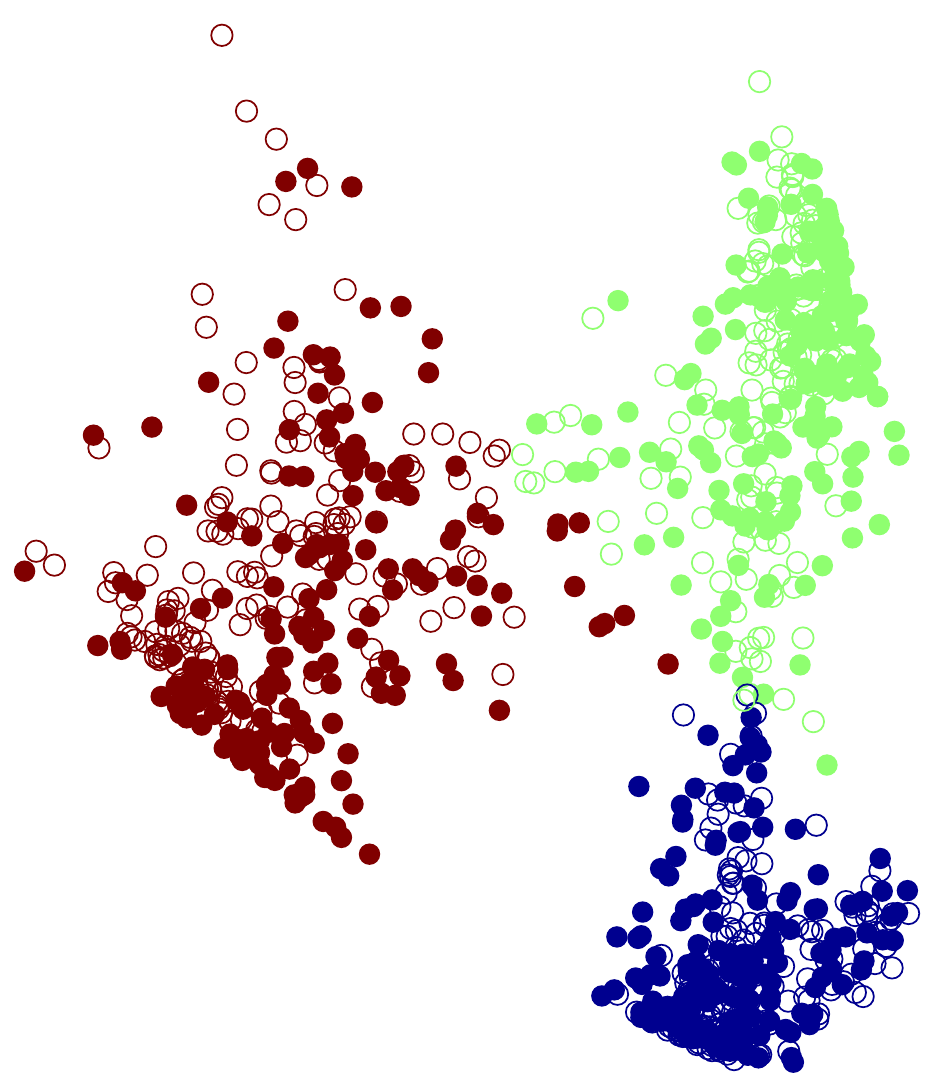}&
\includegraphics[height = 1.55cm]{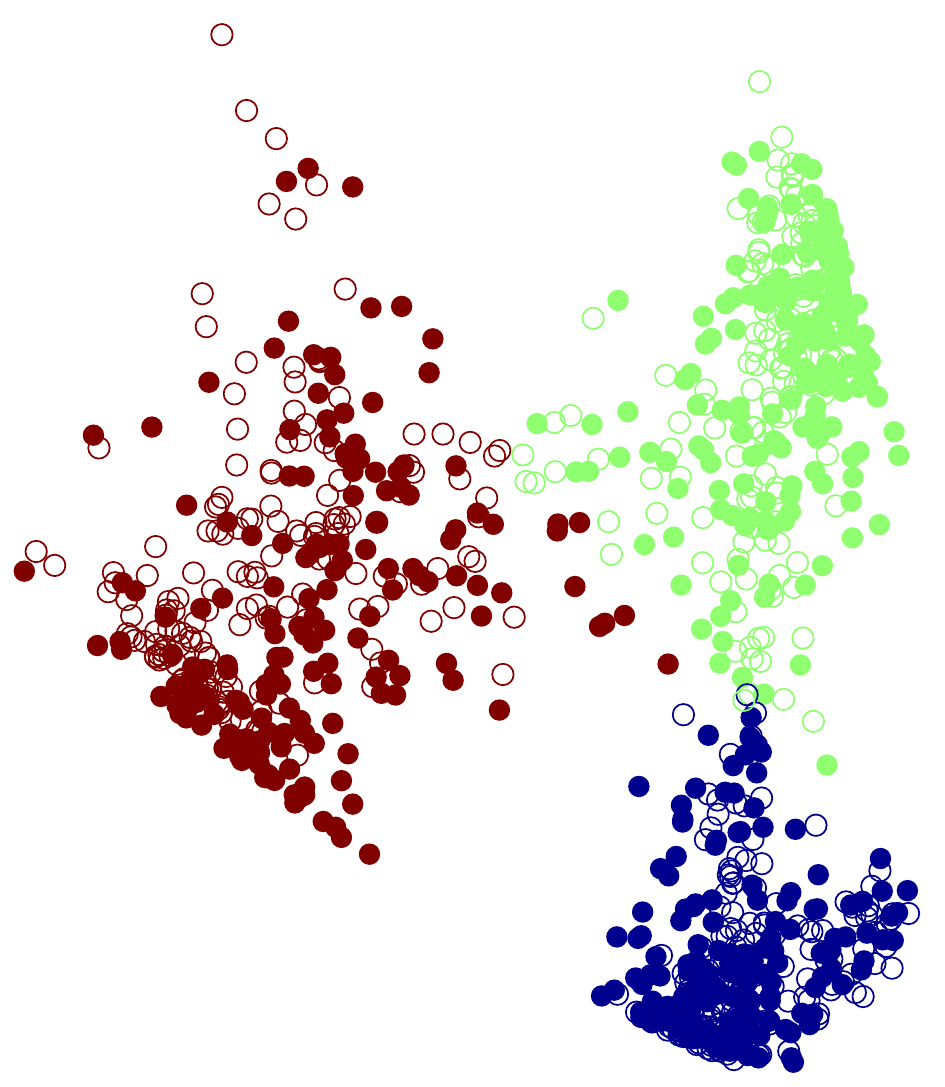}&
\includegraphics[height = 1.55cm]{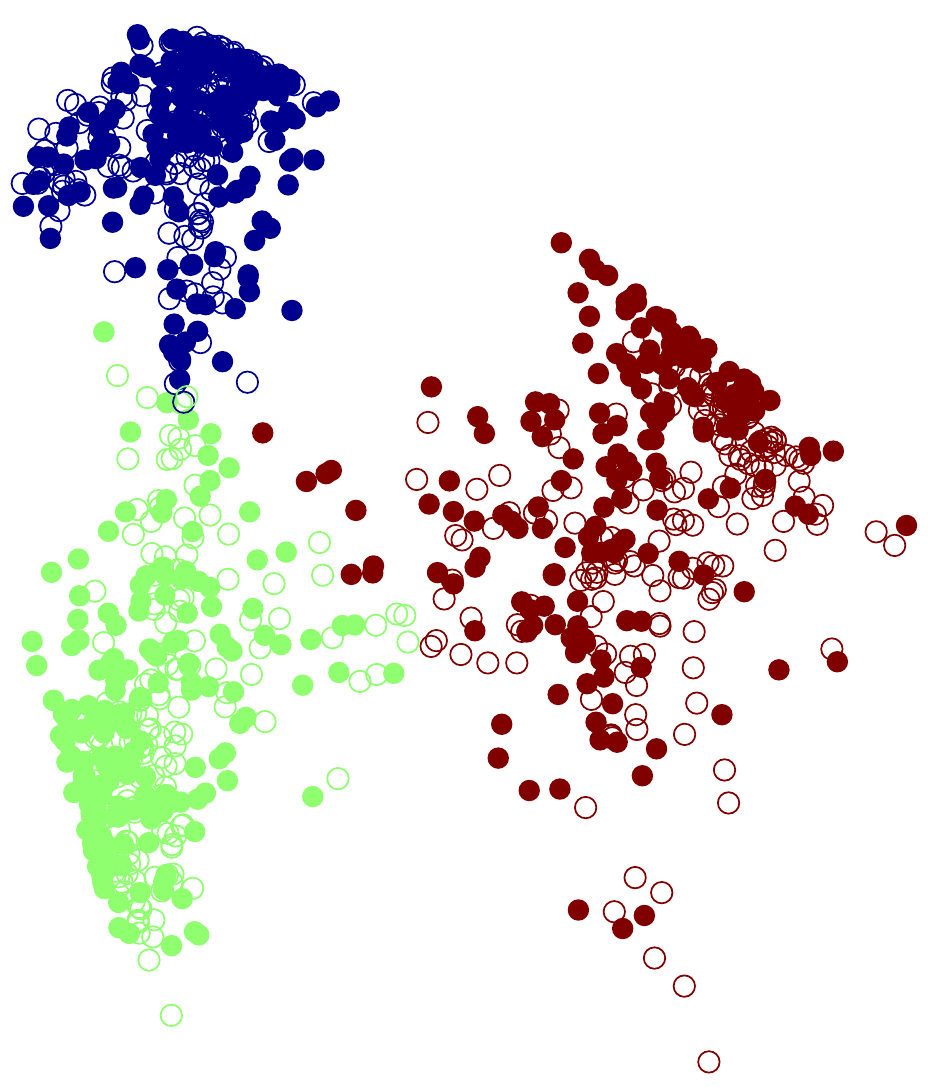}&
\includegraphics[height = 1.55cm]{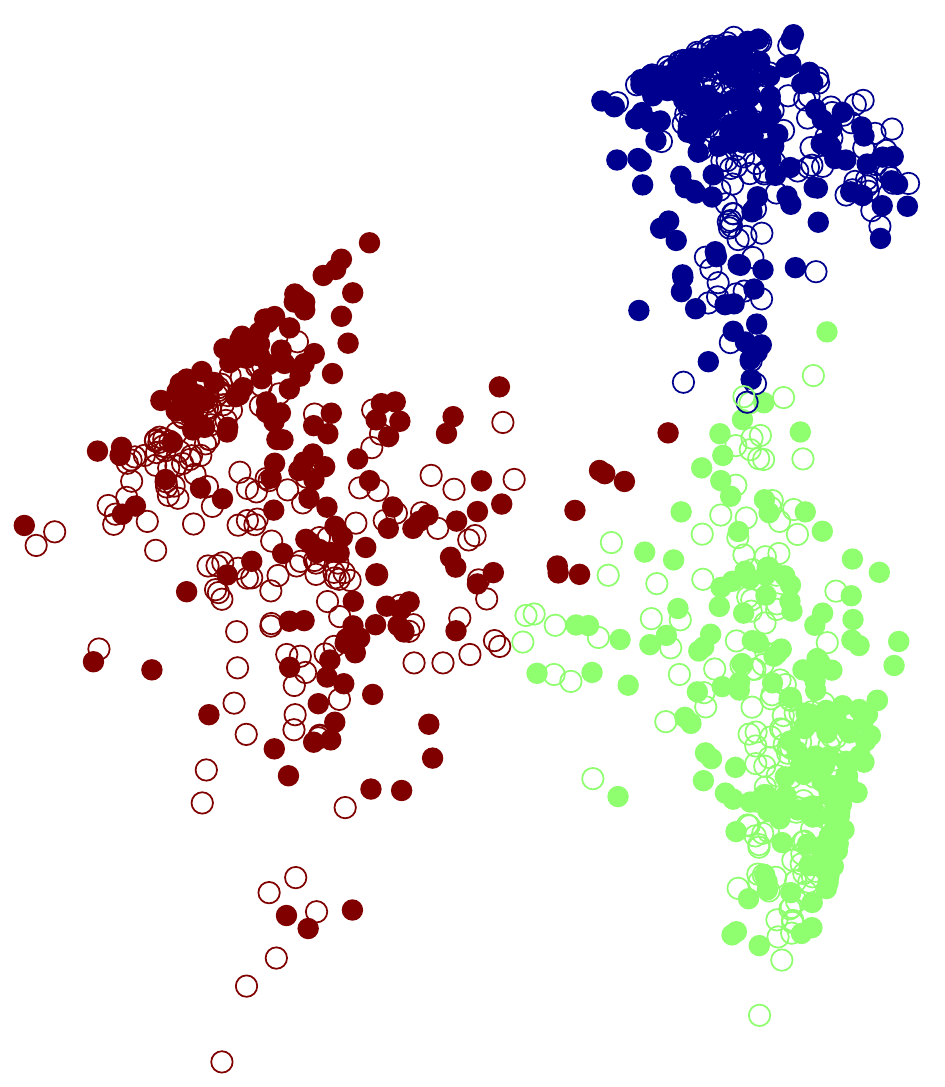}\\\hline
\end{tabular}

\caption{Linear and kernel manifold alignment on the scaled interwined spirals toy experiment (Exp. \#1 in Fig.~\ref{fig:toy}). REKEMA is compared to SSMA for different rates of training samples (we used $l_i=100$ and $u_i=50$ per class for both domains). 
}
\label{fig:sekema}
\end{figure*}

\paragraph{Alignment with REKEMA:} We now consider the reduced-rank approximation of KEMA proposed in Section~\ref{sec:rekema}. We used the data in the experiment \#1 above. Figure~\ref{fig:sekema} illustrates the solutions of the standard SSMA (or KEMA with linear kernel), and for REKEMA using a varying rate of samples. We also give the classification accuracies of a SVM (with both a linear and an RBF kernel) in the projected latent space. Samples were randomly chosen 
 and the sigma parameter for the RBF kernel in KEMA was fixed to the average distance between all used labeled samples. We can observe that SSMA successfully aligns the two domains, but we still need to resort to nonlinear classification to achieve good results. REKEMA, on the contrary, essentially does two operations simultaneously: alignment and data unfolding. Excessive sparsification leads to poor results. Virtually no difference between the full and the reduced-rank solutions are obtained for small values of $r$: just 10\% of examples are actually needed to saturate accuracies.

\paragraph{Invertibility of the projections: }

\begin{figure*}[h!]
\small
\begin{center}
\setlength{\tabcolsep}{0pt}
\begin{tabular}{|c|cc|cc|cc|}
\hline
& \multicolumn{2}{c}{SSMA} & \multicolumn{2}{c}{KEMA ($K_{\text{lin}} \to K_{\text{lin}}$)} & \multicolumn{2}{c|}{KEMA ($K_{\text{RBF}} \to K_{\text{lin}}$)}\\
\cline{2-7}
Exp. & Domains  & Classes & Domains  & Classes & Domains  & Classes \\
\hline\hline
\rotatebox{90}{Exp. \#1}& 
\includegraphics[height = 1.55cm]{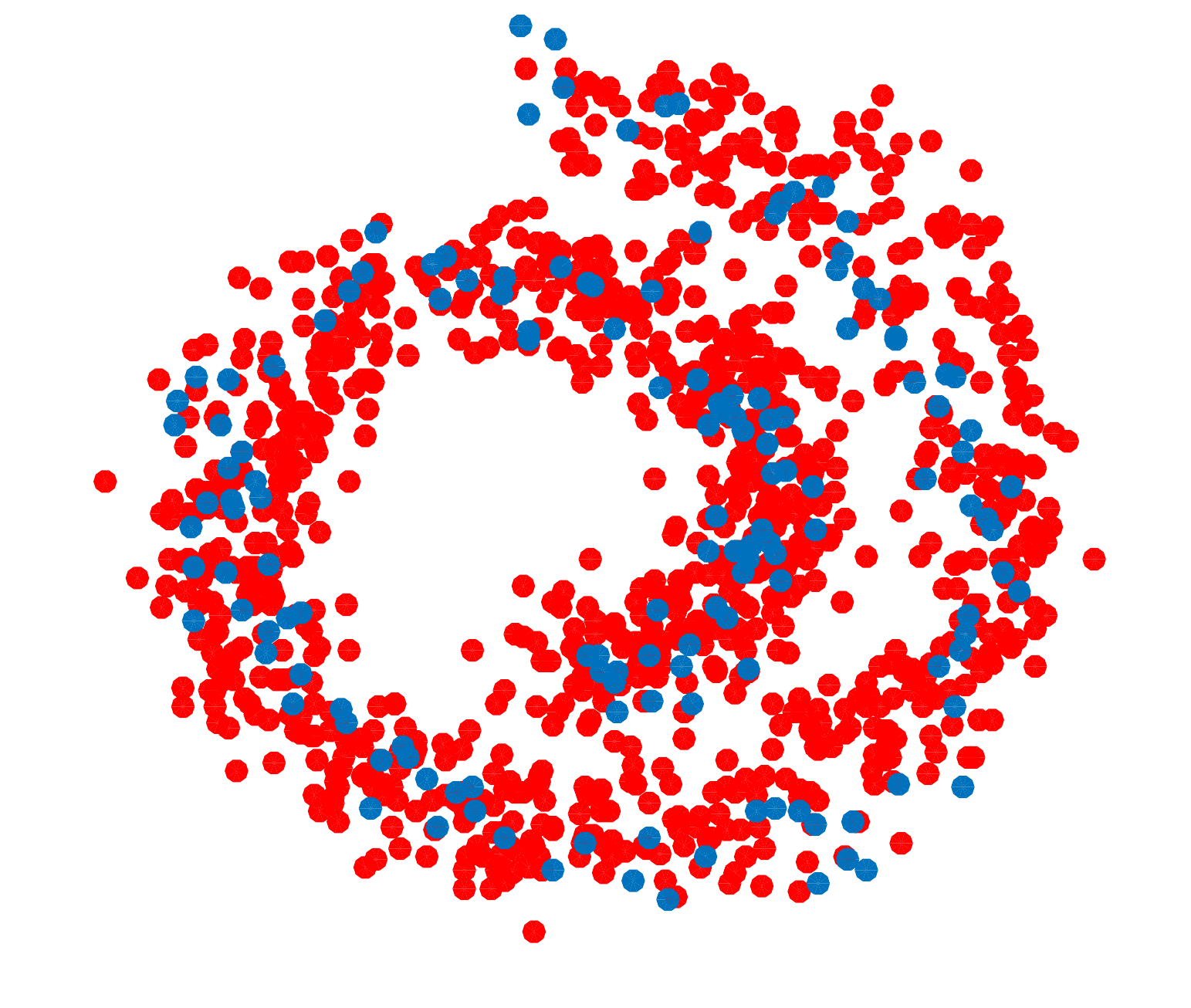}&
\includegraphics[height = 1.55cm]{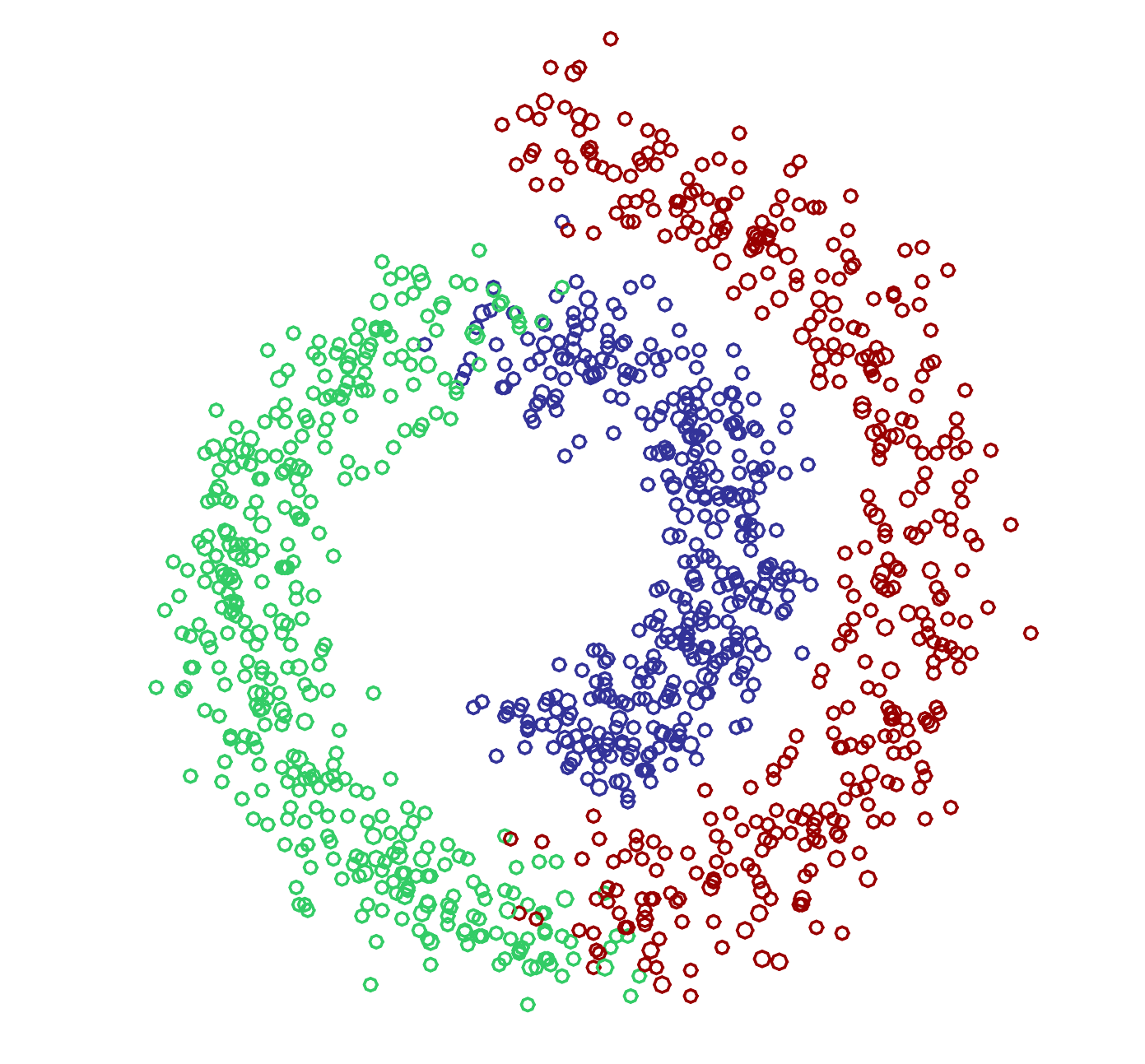}&
\includegraphics[height = 1.55cm]{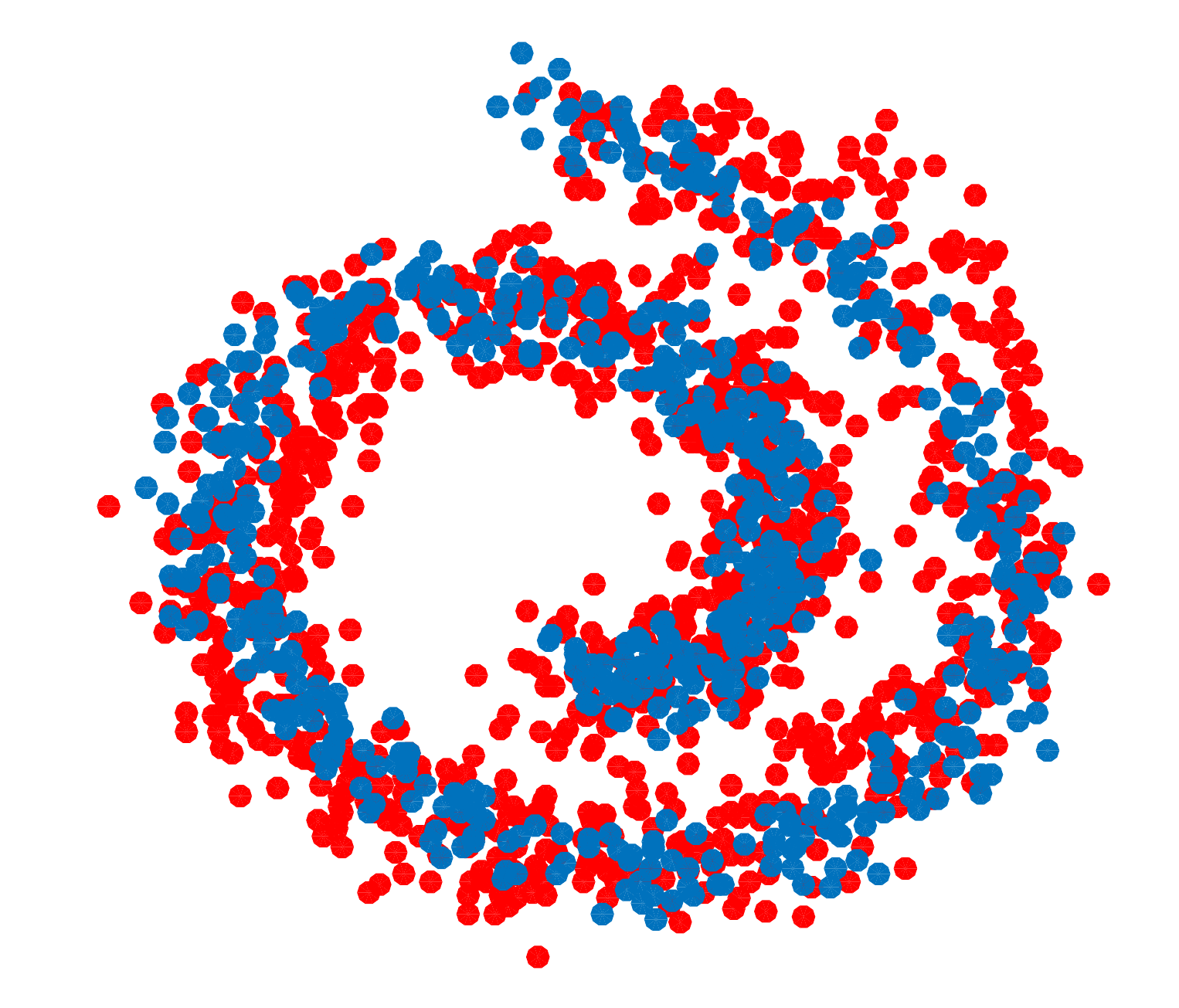}&
\includegraphics[height = 1.55cm]{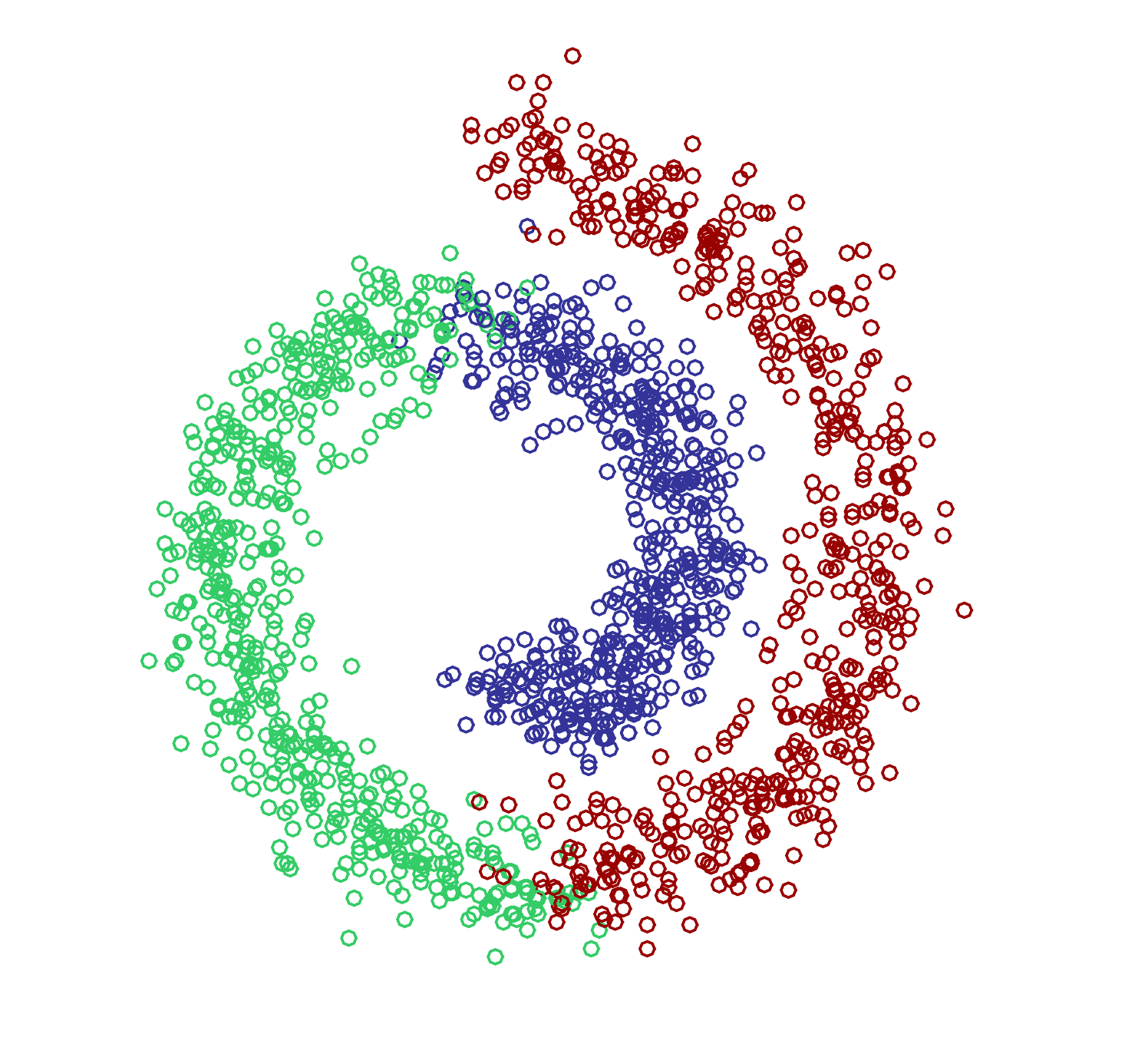}&
\includegraphics[height = 1.55cm]{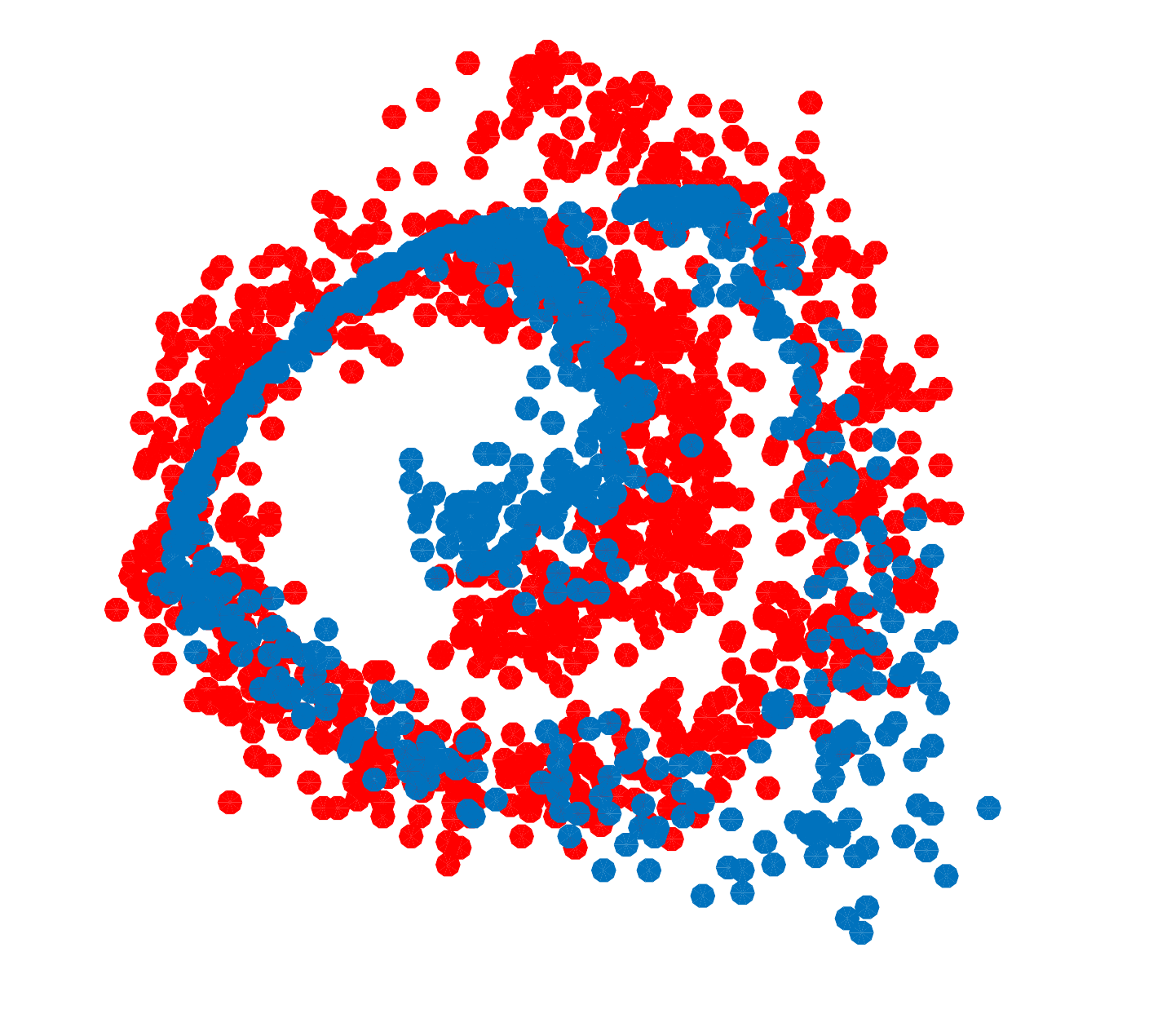}&
\includegraphics[height = 1.55cm]{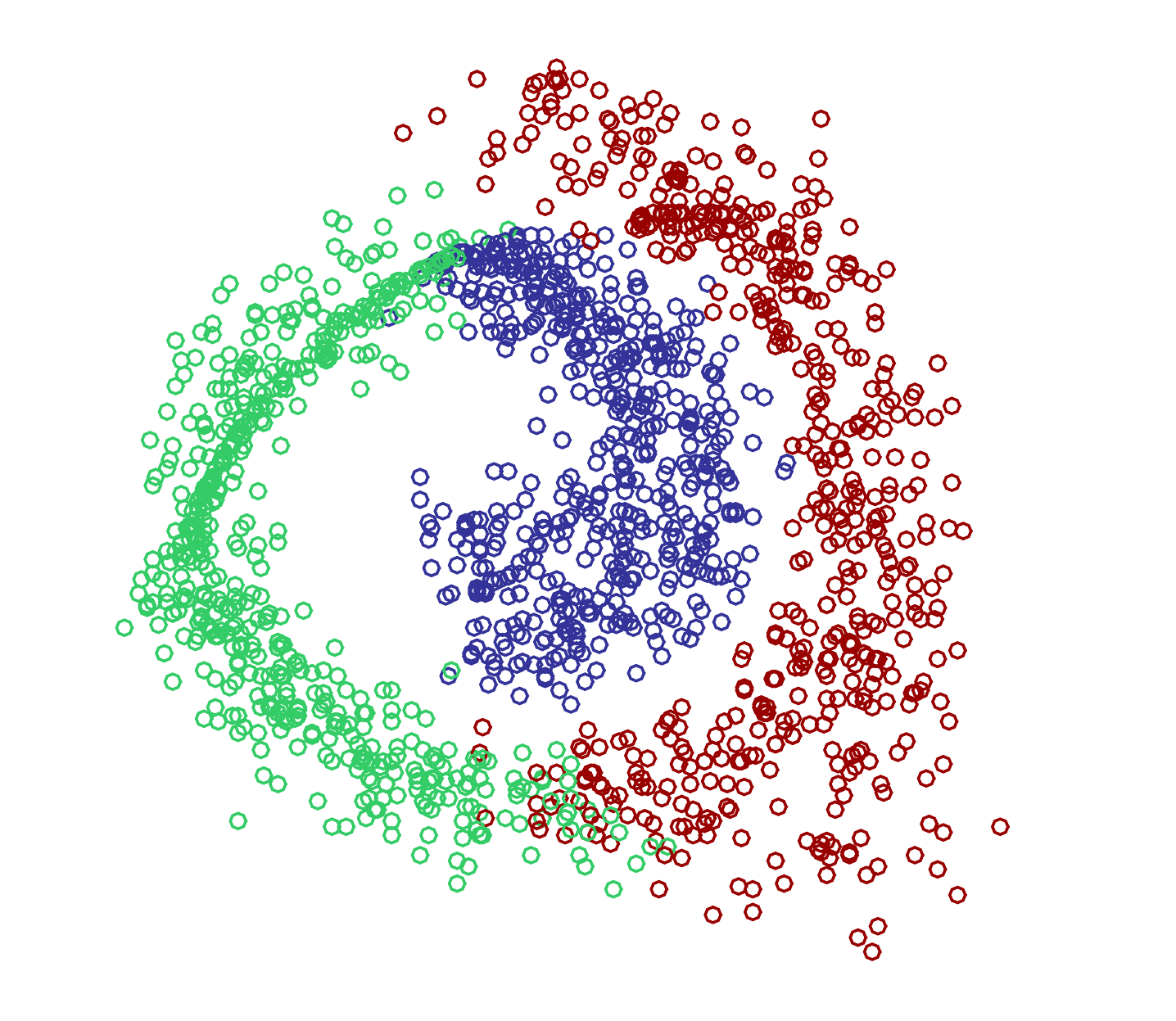}\\
&\multicolumn{2}{c|}{$0.15 \pm 0.01$} & \multicolumn{2}{c|}{$\mathbf{0.14 \pm 0.01}$} & \multicolumn{2}{c|}{$0.23 \pm 0.06$}\\
\hline


\rotatebox{90}{Exp. \#2}& 
\includegraphics[height = 1.55cm]{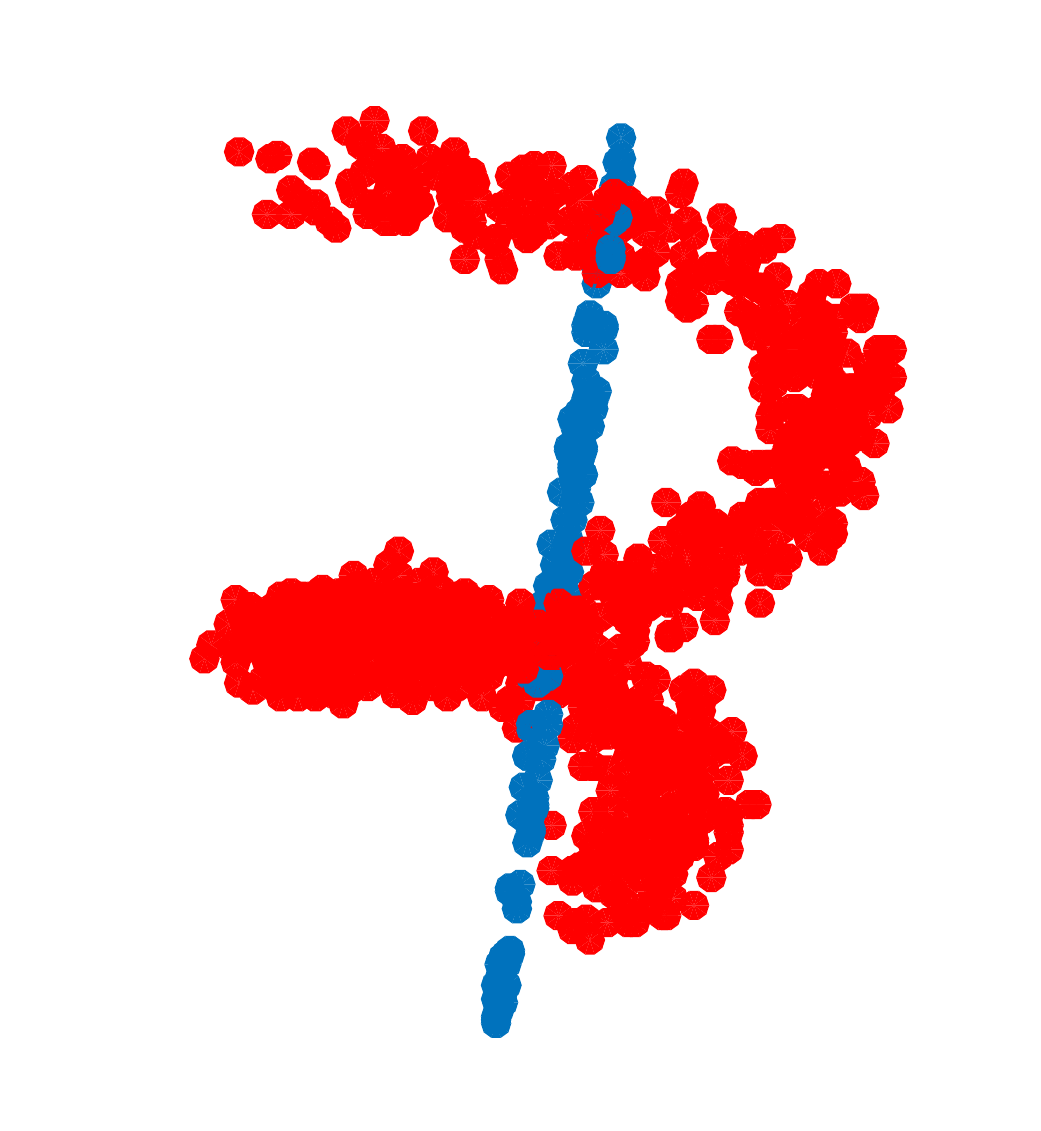}&
\includegraphics[height = 1.55cm]{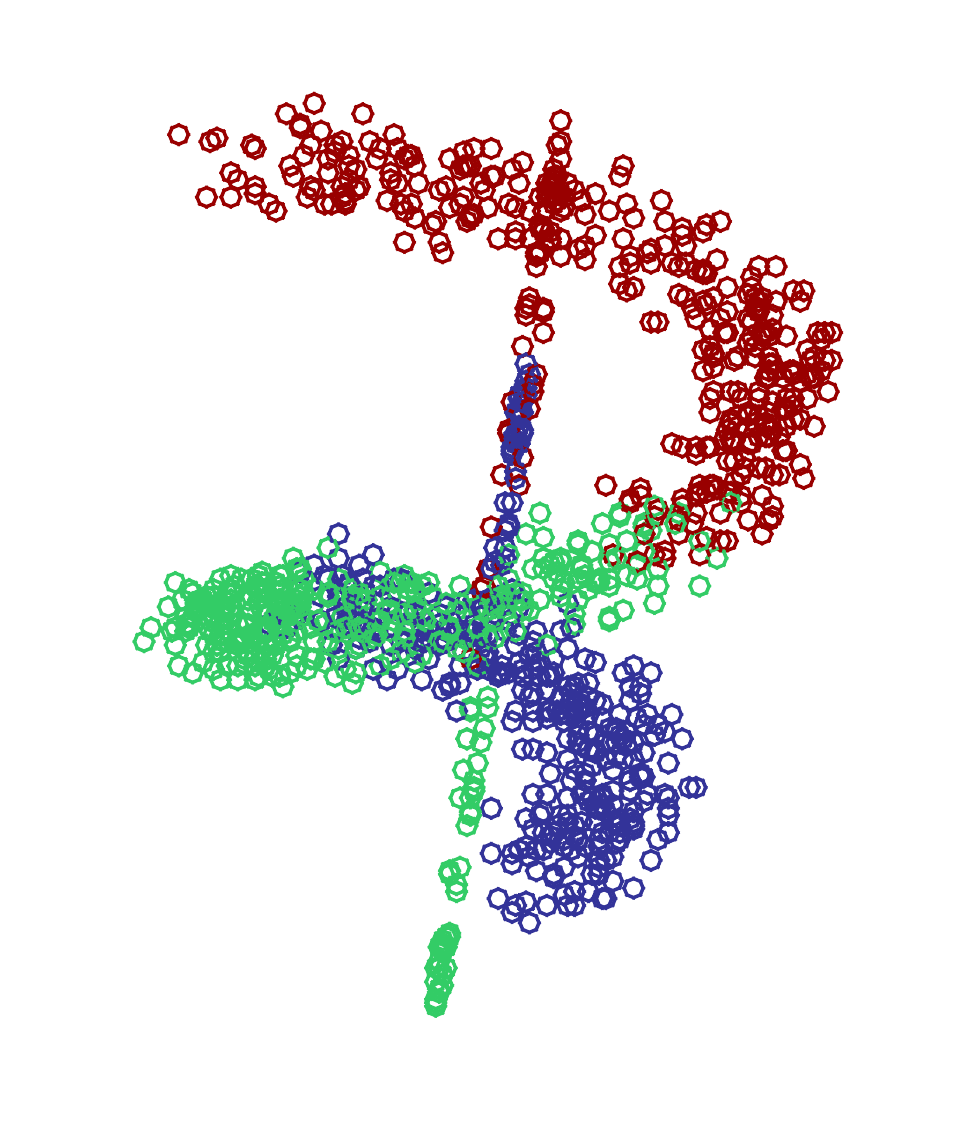}&
\includegraphics[height = 1.55cm]{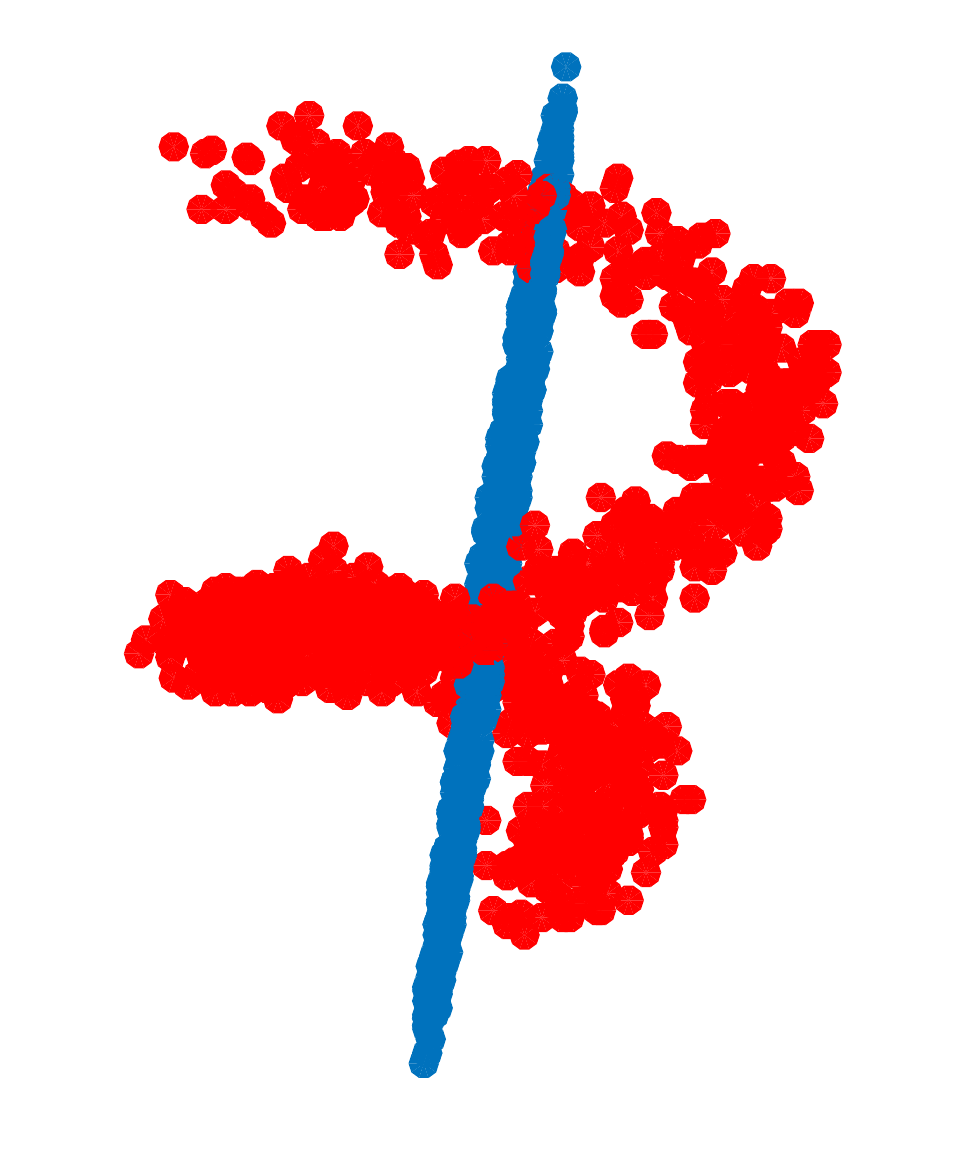}&
\includegraphics[height = 1.55cm]{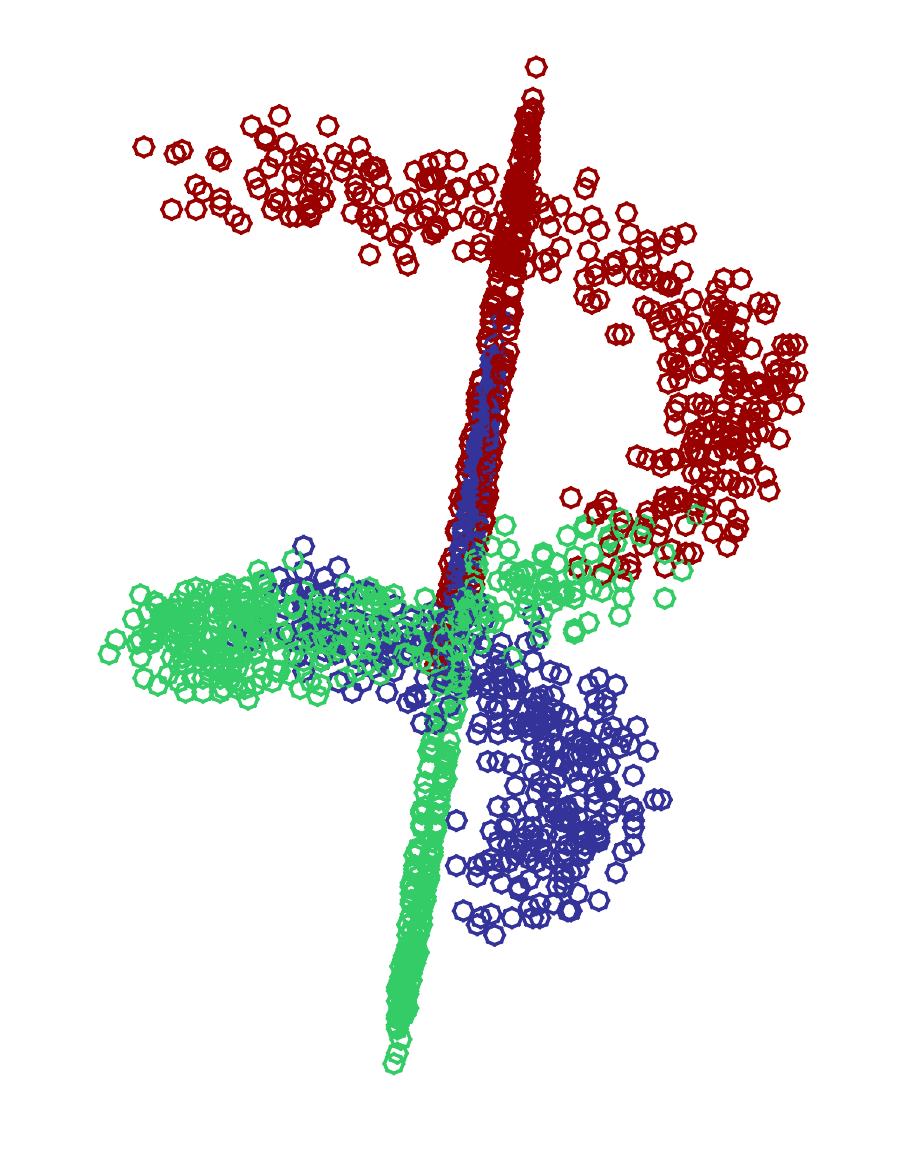}&
\includegraphics[height = 1.55cm]{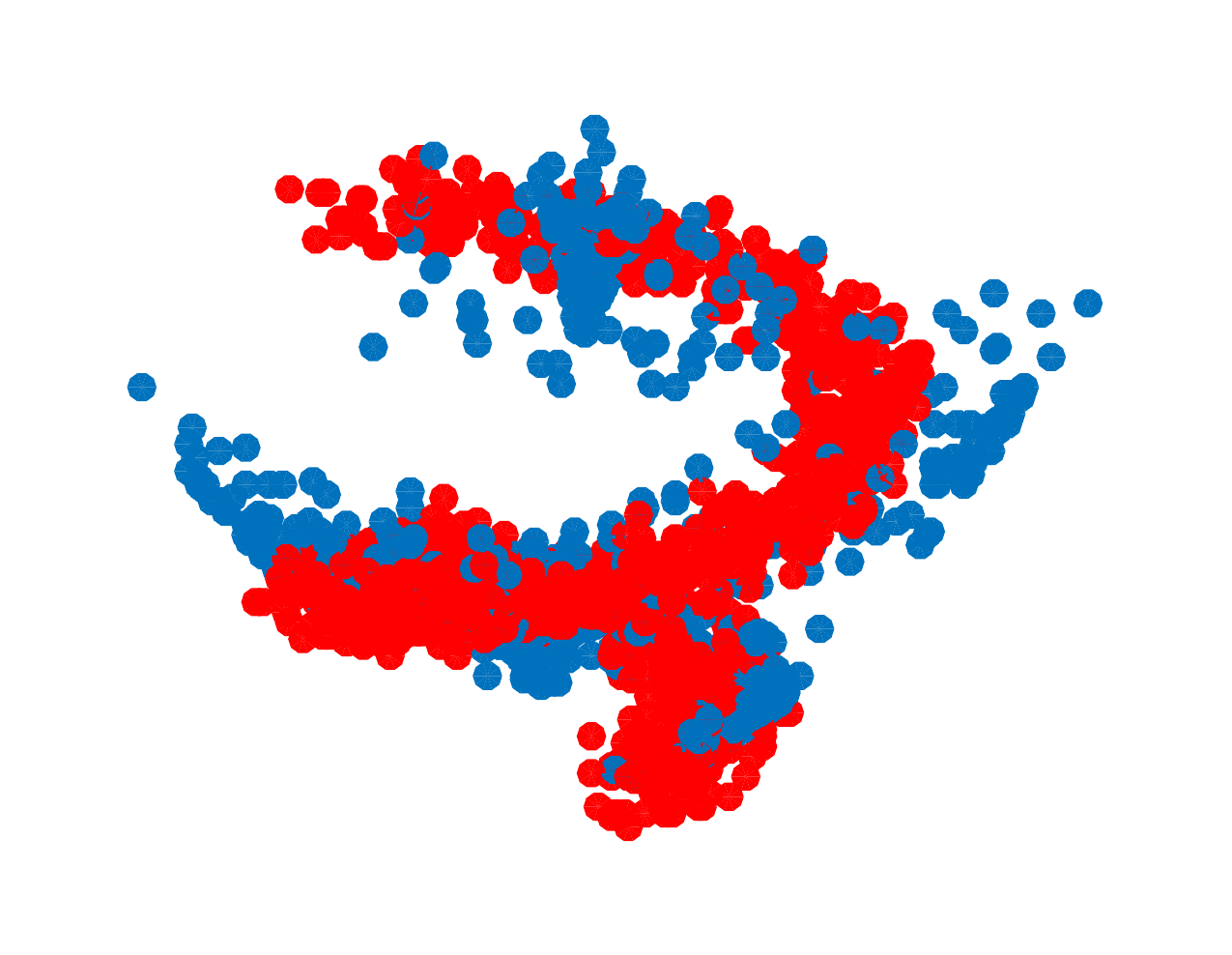}&
\includegraphics[height = 1.55cm]{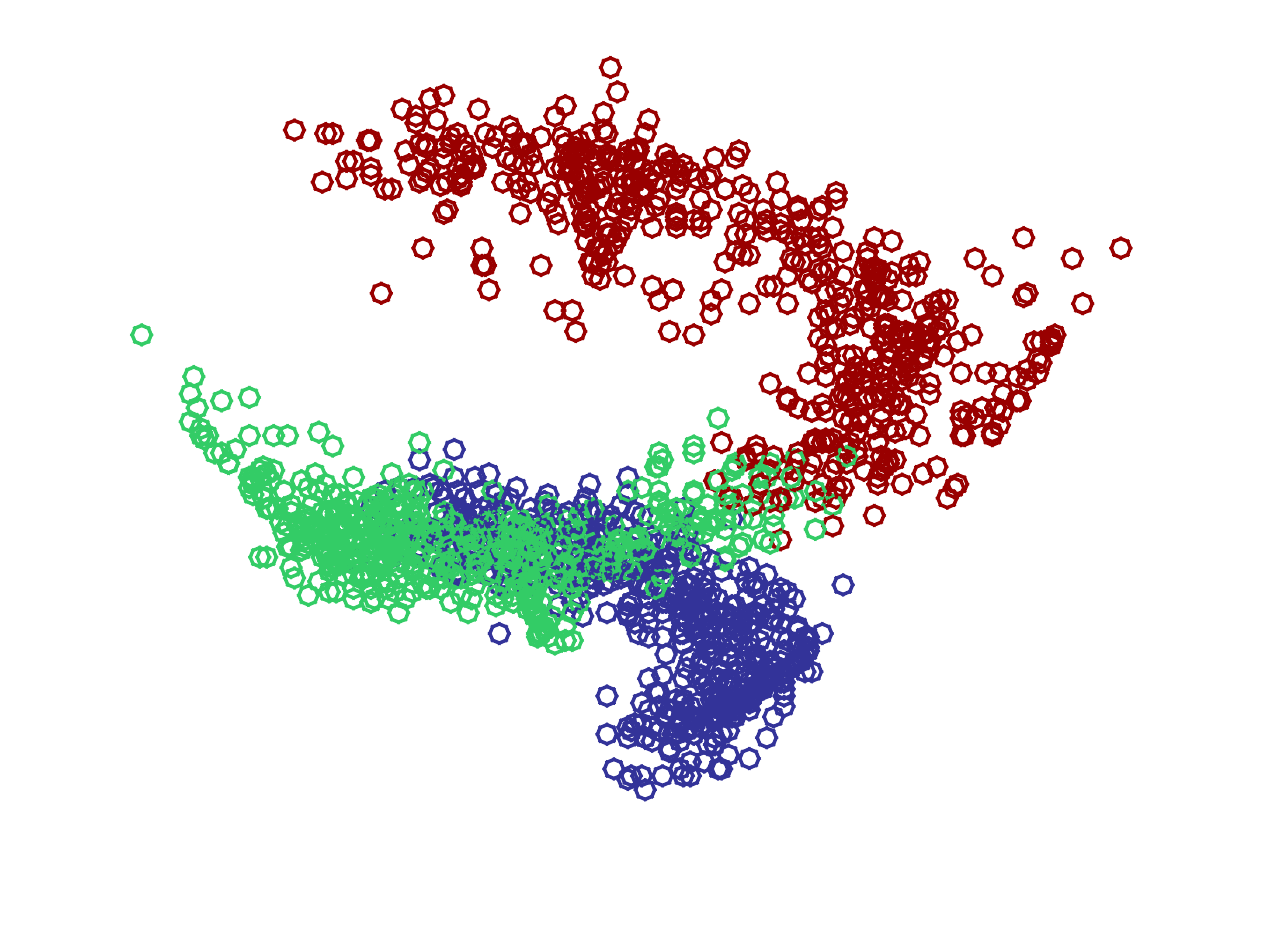}\\
&\multicolumn{2}{c|}{$0.86 \pm 0.01$} & \multicolumn{2}{c|}{$0.87 \pm 0.01$ } & \multicolumn{2}{c|}{$\mathbf{0.34 \pm 0.01}$}\\
\hline

\rotatebox{90}{Exp. \#3}& 
\includegraphics[height = 1.55cm]{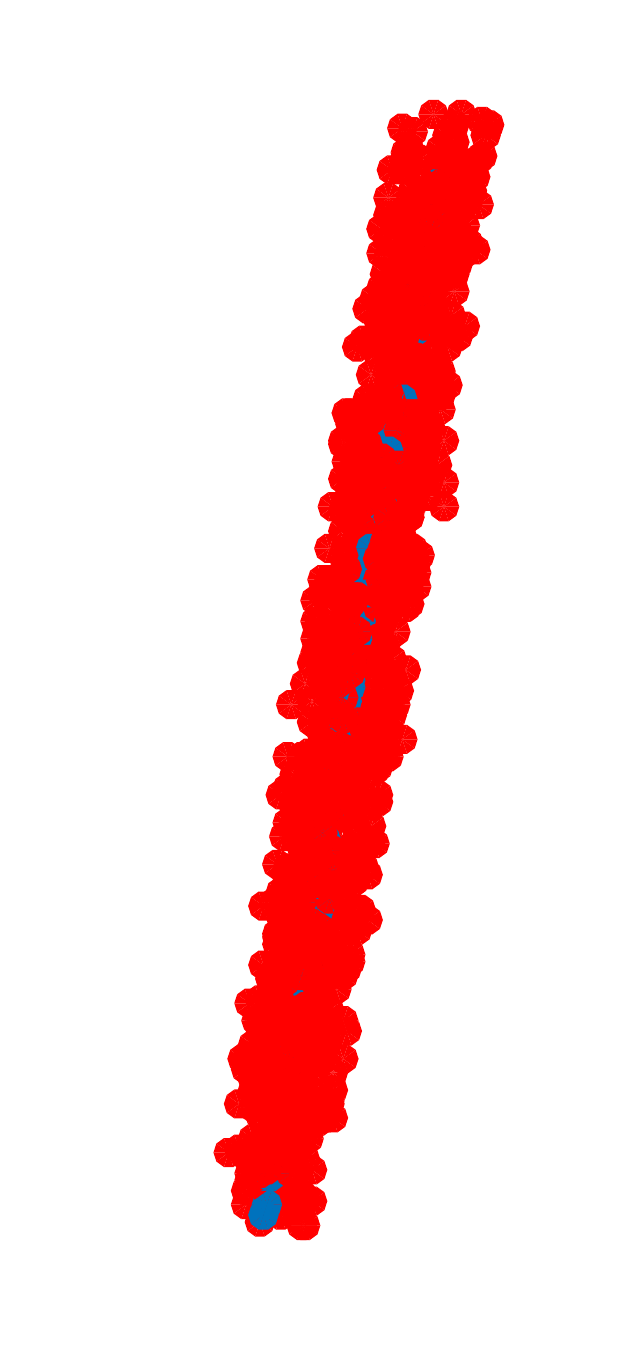}&
\includegraphics[height = 1.55cm]{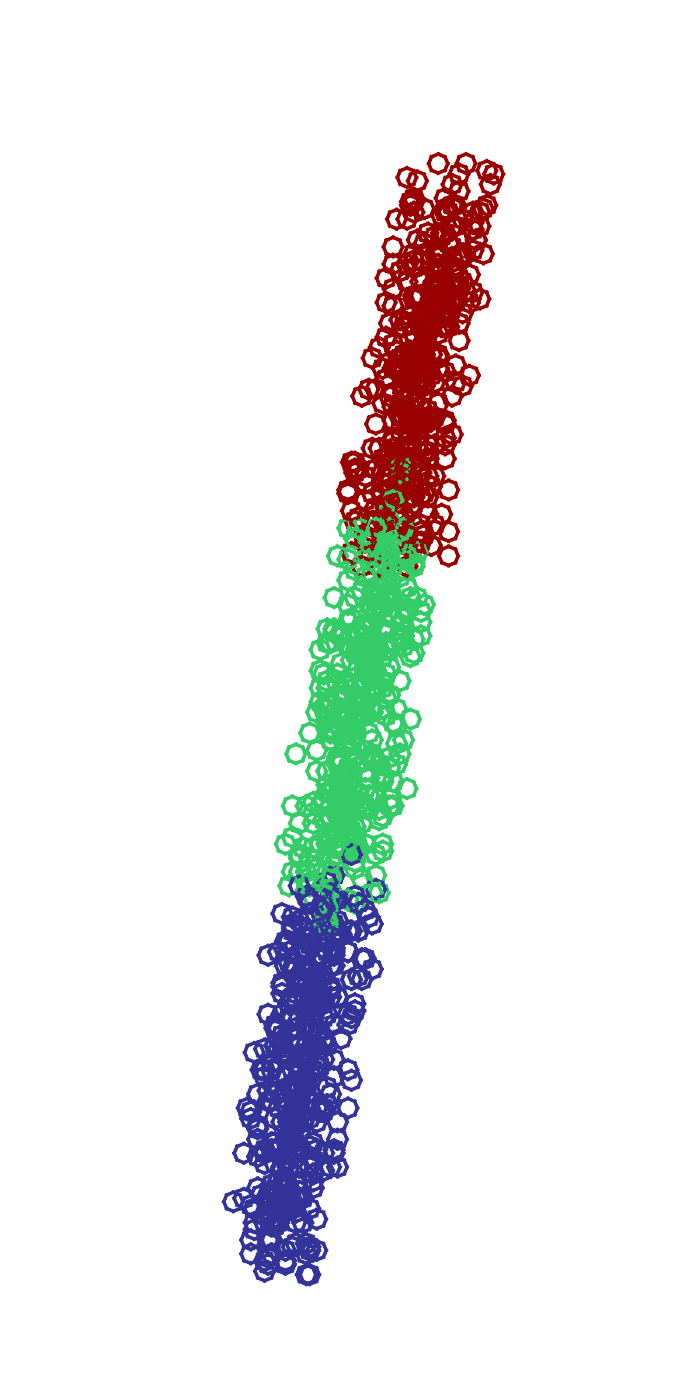}&
\includegraphics[height = 1.55cm]{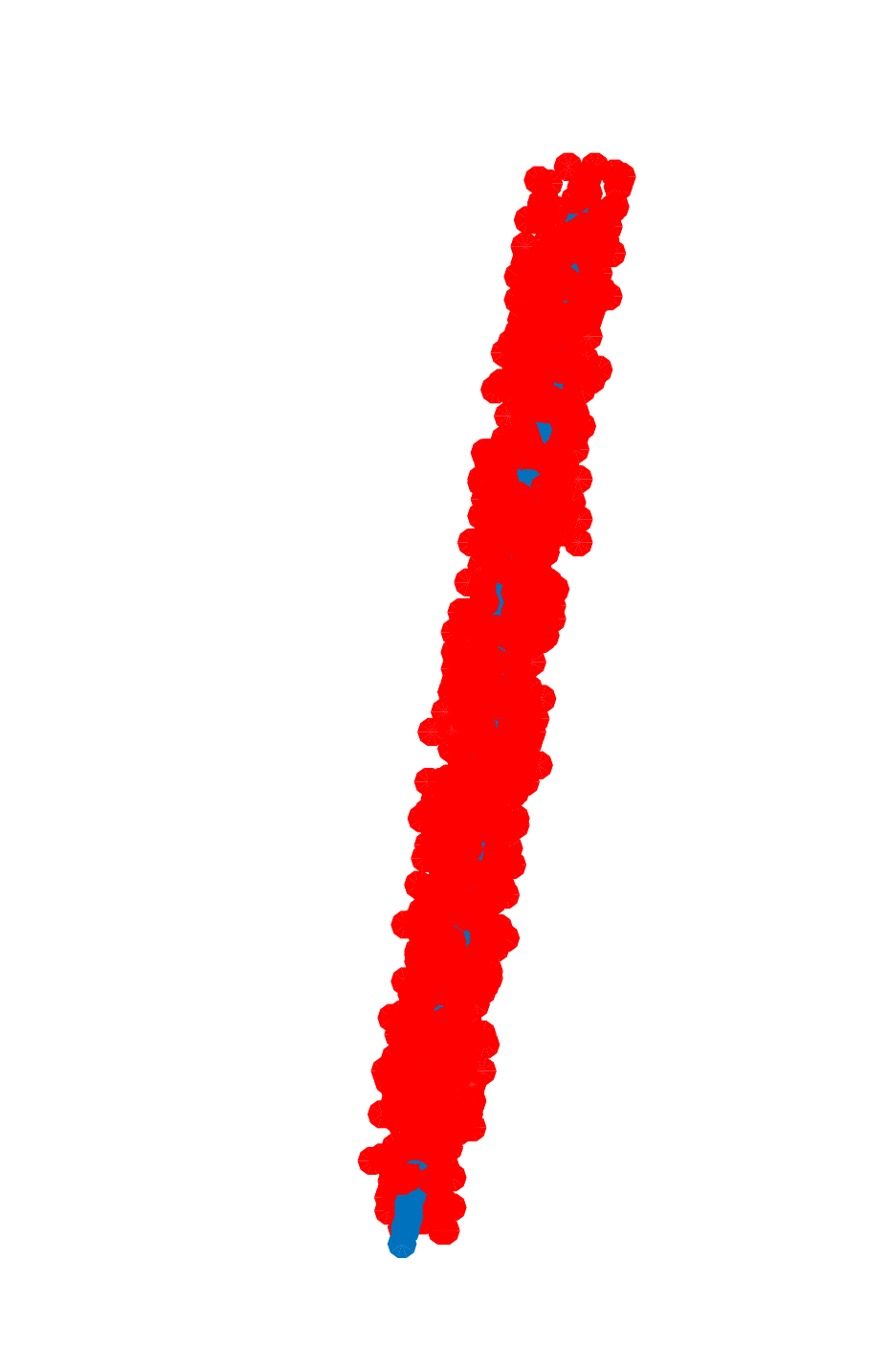}&
\includegraphics[height = 1.55cm]{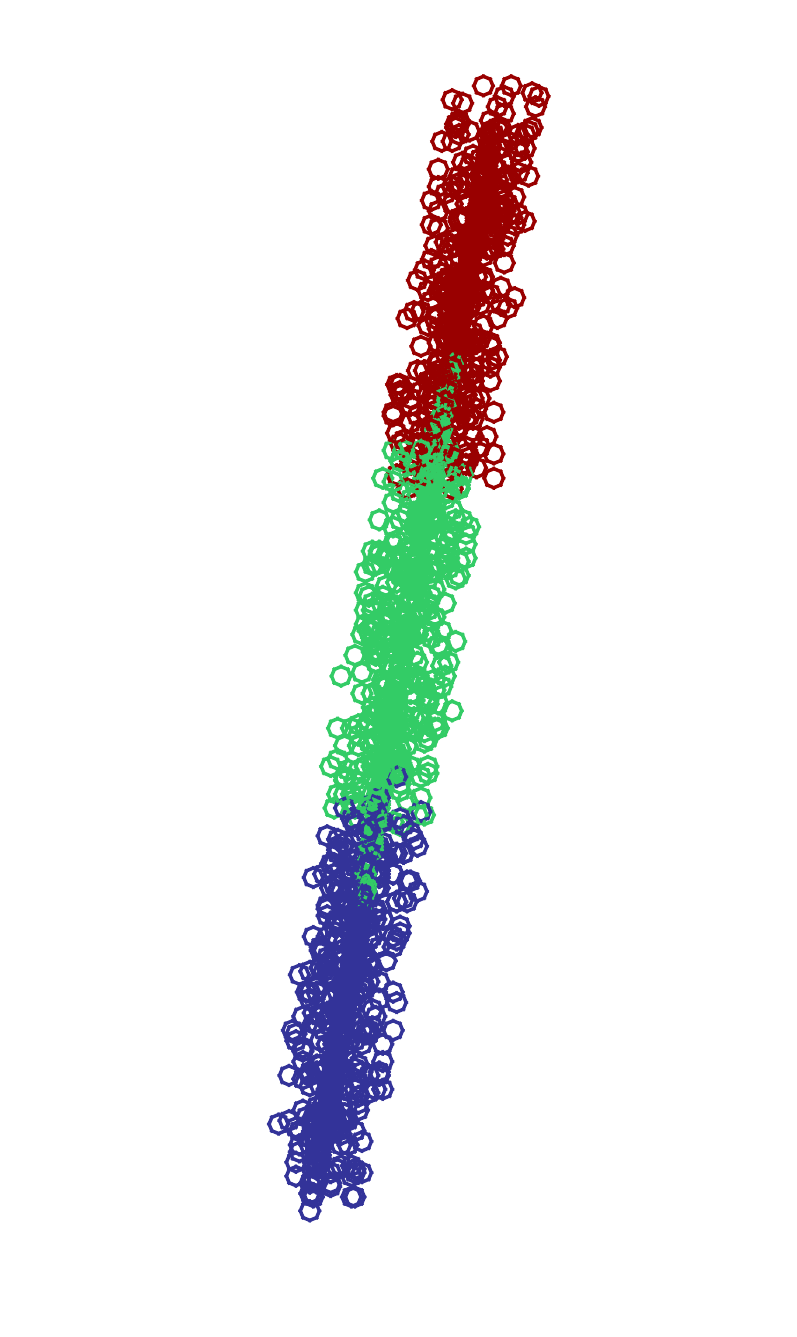}&
\includegraphics[height = 1.55cm]{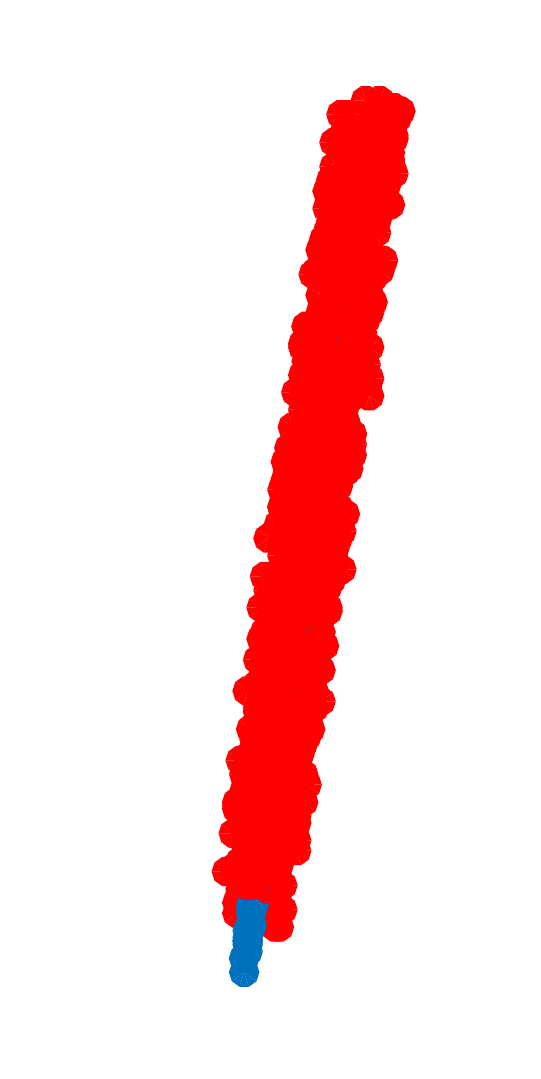}&
\includegraphics[height = 1.55cm]{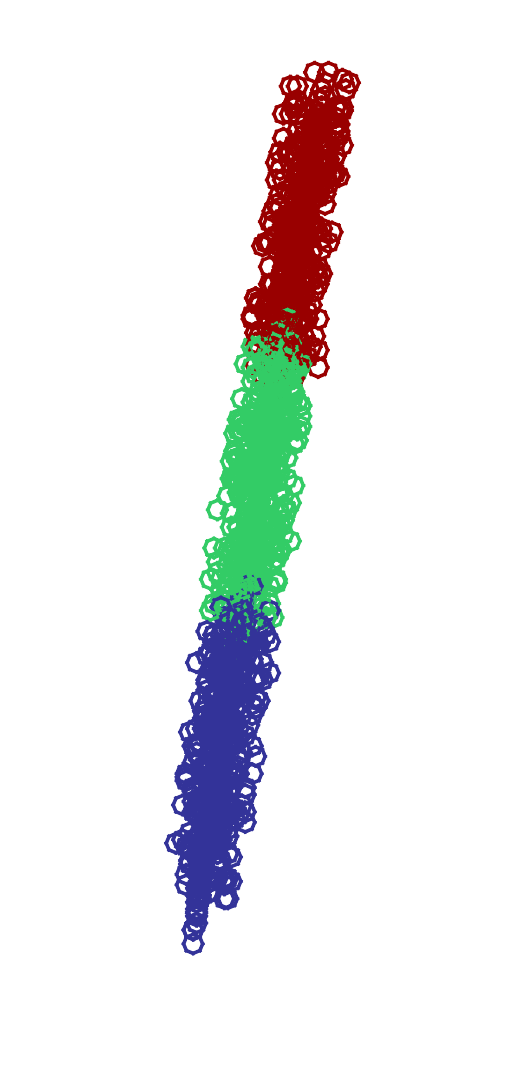}\\
&\multicolumn{2}{c|}{$0.19 \pm 0.05$} & \multicolumn{2}{c|}{$\mathbf{0.15 \pm 0.01}$ } & \multicolumn{2}{c|}{$0.16 \pm 0.01$}\\
\hline

\rotatebox{90}{Exp. \#4}& 
\includegraphics[height = 1.55cm]{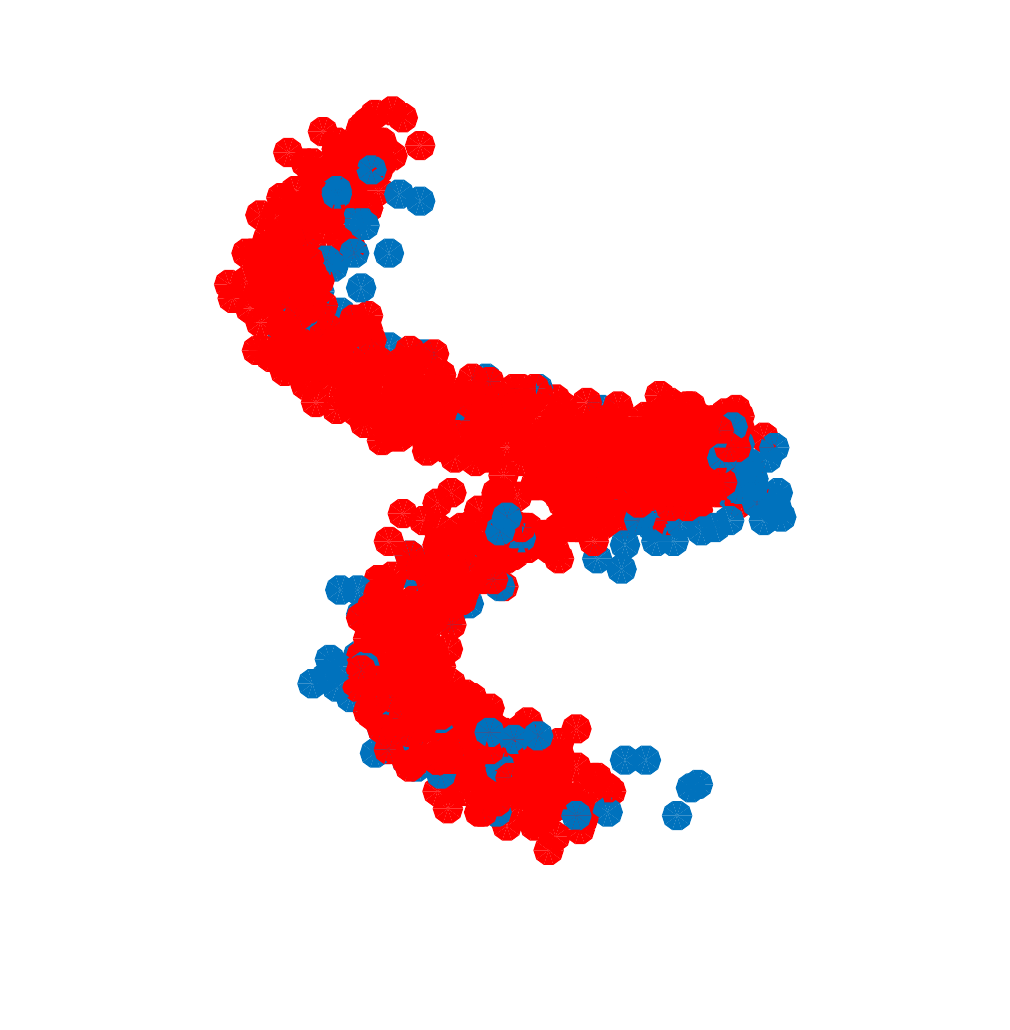}&   
\includegraphics[height = 1.55cm]{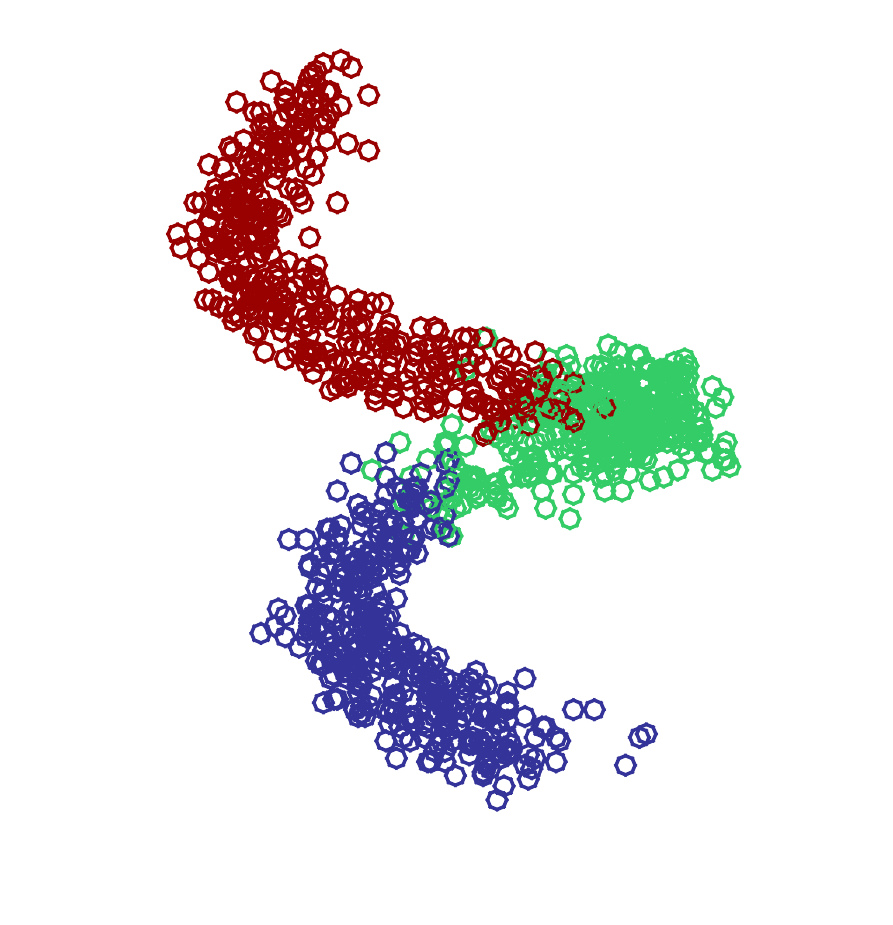}&   
\includegraphics[height = 1.55cm]{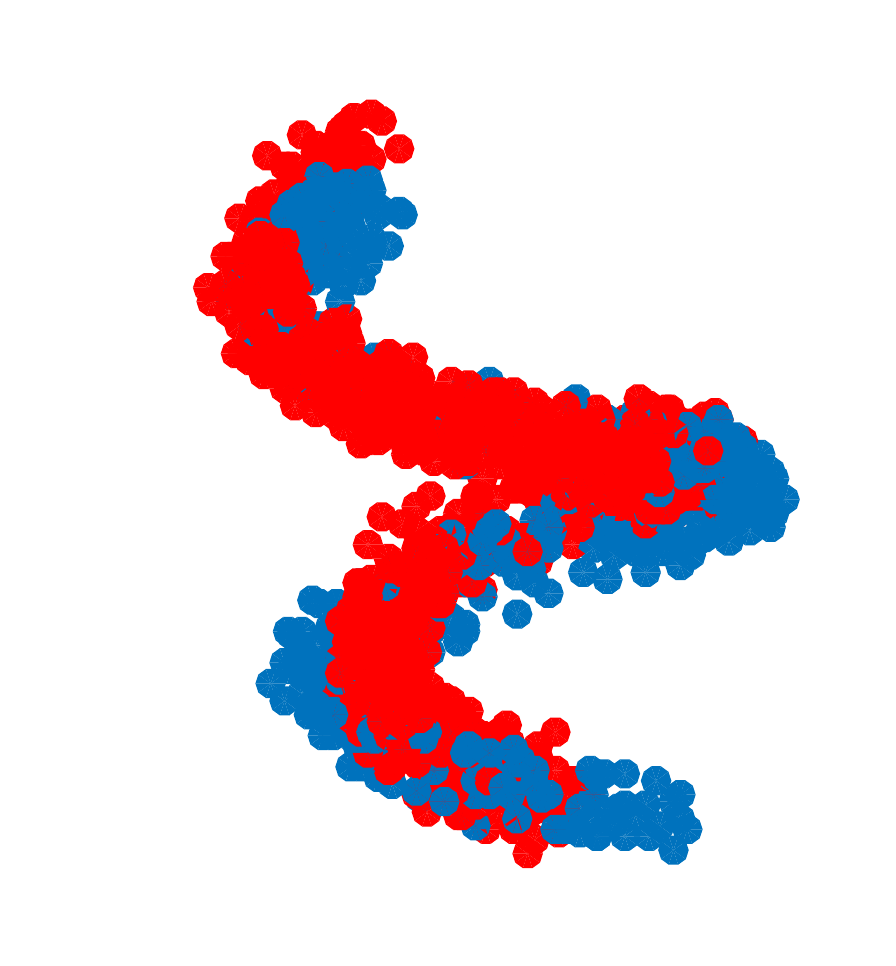}&
\includegraphics[height = 1.55cm]{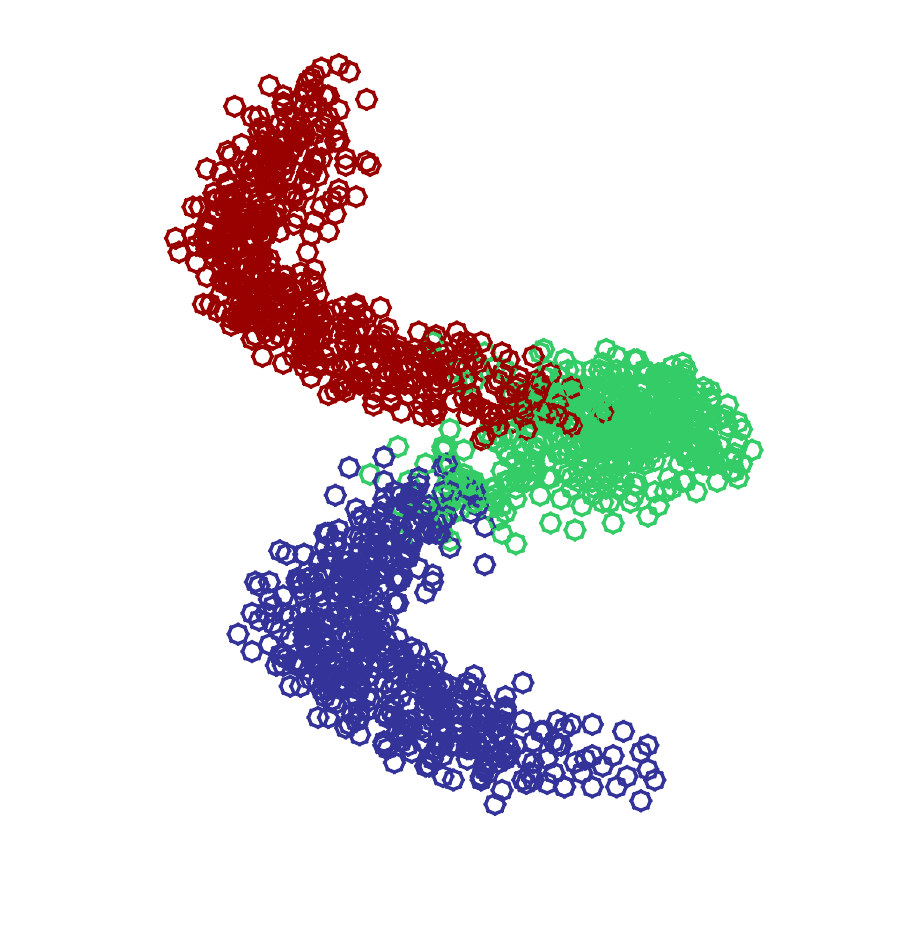}&
\includegraphics[height = 1.55cm]{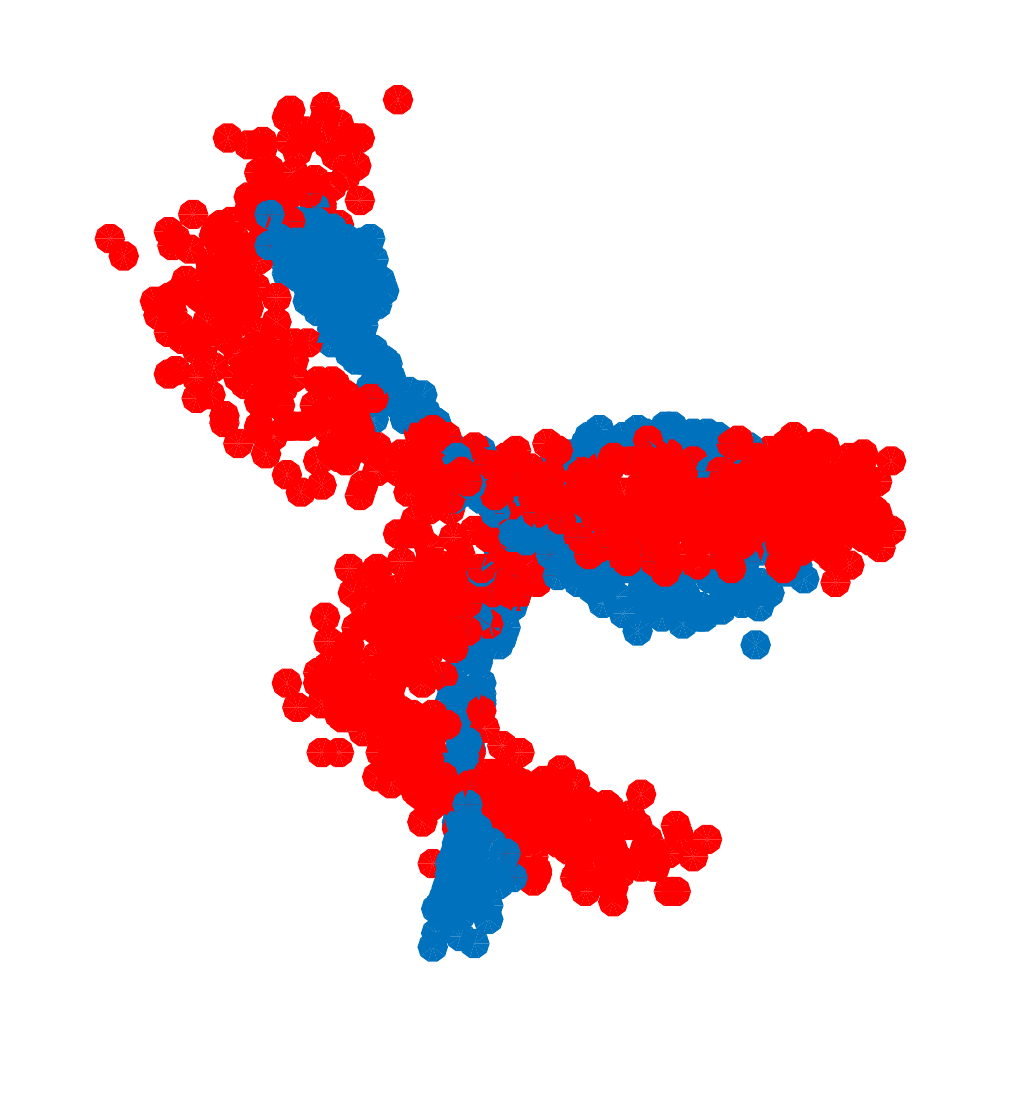}&
\includegraphics[height = 1.55cm]{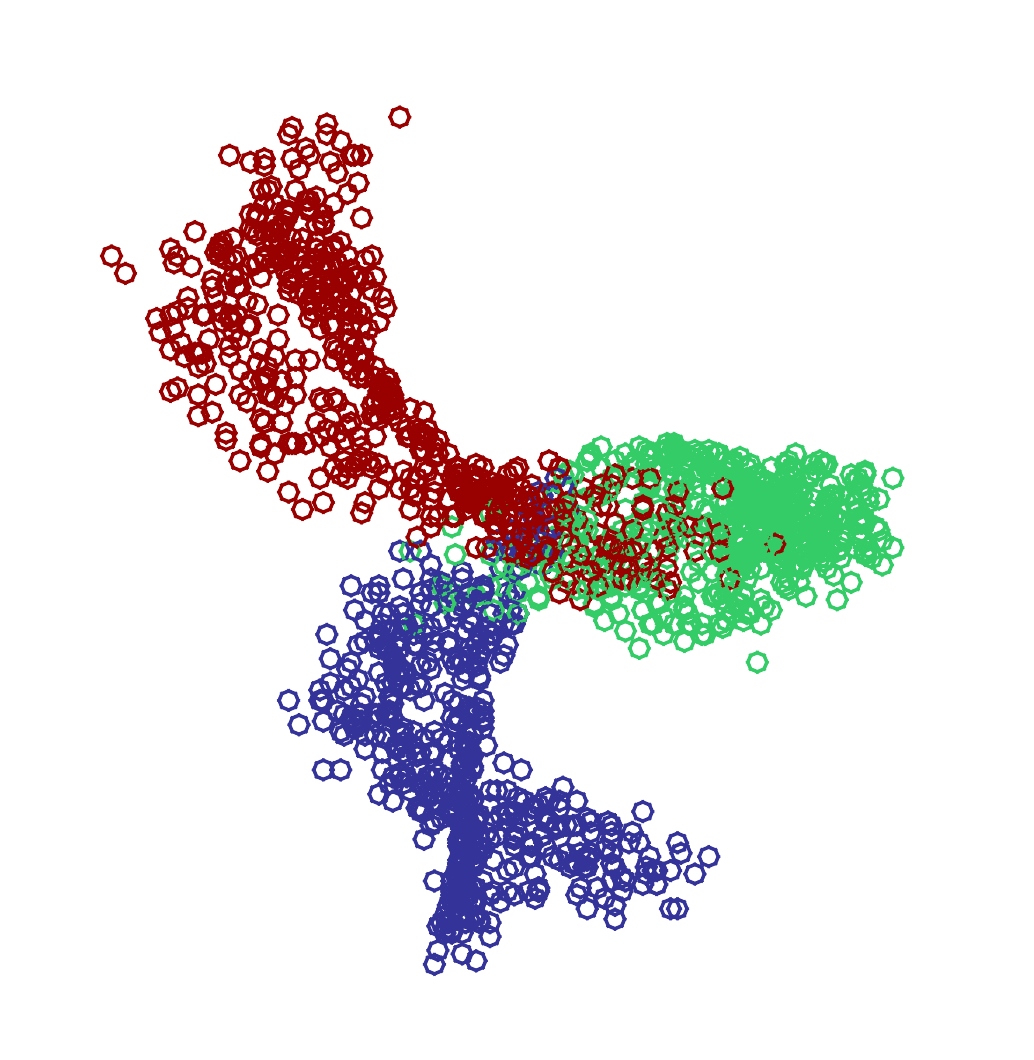}\\
&\multicolumn{2}{c|}{$0.43 \pm 0.03$} & \multicolumn{2}{c|}{$0.43 \pm 0.03$ } & \multicolumn{2}{c|}{$\mathbf{0.42 \pm 0.04}$}\\
\hline


\end{tabular}
\caption{Domain inversion with SSMA and KEMA. ($\red{\bullet}$) = samples in the source domain, ($\blue{\bullet}$) =  target domain samples projected onto the source domain, and ($\red{\bullet}$, $\green{\bullet}$, $\blue{\bullet}$) =  class distributions. Each plot shows the result of a single run, and the averaged $\ell_2$-norm reconstruction error over 10 runs.}
\label{fig:inv}
\end{center}
\end{figure*}

Figure~\ref{fig:inv} shows the results of invertibility of SSMA and KEMA (using Eq.~\eqref{eq:inv}) on the previous toy examples. We use a linear kernel for the inversion part (latent-to-source) and use for the direct part (target-to-latent space)  either a linear or an RBF kernel. All results are shown in the source domain space.  All the other settings (\# labeled and unlabeled, $\mu$, graphs) are kept as in the experiments shown in Fig.~\ref{fig:toy}. The reconstruction error, averaged on 10 runs, is also reported: 
KEMA is capable of inverting the projections and is always as accurate as the SSMA method in the simplest cases (\#1, \#4). For the cases related to higher levels of deformation, KEMA is either as accurate as SSMA (\#3, where the inversion is basically a projection on a line) or significantly better: e.g. for experiment \#2, where the two domain are strongly deformed, only KEMA with RBF kernel can achieve satisfying inversion, as it unfolds the target domain and then only needs a rotation to match the distribution in the source domain. 


\subsection{Visual object recognition in multi-modal datasets}

We here evaluate KEMA on visual object recognition tasks by using the dataset introduced in~\cite{Saenko10}. We consider the four domains Webcam (\texttt{W}), Caltech (\texttt{C}), Amazon (\texttt{A}) and DSLR (\texttt{D}), and selected the 10 common classes in the four datasets following~\cite{gong12}. By doing so, the domains contain 295 (Webcam), 1123 (Caltech), 958 (Amazon) and 157 (DSLR) images, respectively. The features were extracted as described in~\cite{Saenko10}: we use a $800$-dimensional normalized histogram of visual words obtained from a codebook constructed from a subset of the Amazon dataset on points of interest detected by the Speeded Up Robust Features (SURF) method. We used the same experimental setting as~\cite{gopalan11,gong12}, in order to compare with these unsupervised domain adaptation methods. Additionally, we compare our proposal with the following semi-supervised domain adaptation methods: SGF~\cite{gopalan11}, GFK~\cite{gong12}, OT-lab~\cite{Cou14} and MMDT~\cite{Hof13}.

For all methods, we used $20$ labeled pixels {\em per} class in the source domain for the \texttt{C}, \texttt{A} and \texttt{D} domains and 8 samples {\em per} class for the \texttt{W} domain. After alignment, an ordinary $1$-NN classifier was trained with the labeled samples. The same labeled samples in the source domain were used to define the PLS eigenvectors for GFK and OT-lab. For all the methods using labeled samples in the target domain (including KEMA), we used $3$ labeled samples in target domain to define the projections. 

We used sensible kernels for this problem in KEMA: the (fast) histogram intersection kernel, $K_{i}(\x_j,\x_k) = \sum_{d} \min\{x_j^d,x_k^d\}$, and the $\chi_2$ kernel, $K_{\chi_2}(\x_j,\x_k) = \exp(-\chi^2/(2\sigma^2))$, with $\chi^2 = \frac{1}{2} \sum_d (x_j^d-x_k^d)^2/(x_j^d+x_k^d)$~\cite{Sre10}. 
We used $u = 300$ unlabeled samples to compute the graph Laplacians, for which a $k$-NN graph with $k=21$ was used. 
The numerical results obtained in all the eight problems are reported in Table~\ref{tab:Web}: KEMA outperforms all the unsupervised competing methods and, in most of the cases, improves the results obtained by the semi-supervised methods using labels in the source domain only. 
KEMA provides the most accurate results in 3 out of the 8 settings when confronted to state-of-the-art (semi)-supervised algorithms, and similar peformance to state-of-the-art GFK in 6 out of the 8 settings. KEMA is as accurate as the state of the art, but with the advantage of handling naturally domains of different dimensionality, and not requiring a discriminative classifier to align the domains such as for MMDT.

\begin{table*}[!h]
\scriptsize
\begin{center}
\caption{Accuracy in the visual object recognition study (\texttt{C}: Caltech, \texttt{A}: Amazon, \texttt{D}: DSLR, \texttt{W}: Webcam). $1$-NN classification testing on all samples from the target domain ($l_{\text{domain}}$: number of labels per class, $^*$ = results reported in~\cite{Cou14}, $^\dagger$ = results reported in~\cite{Hof13}).}
\label{tab:Web}
\setlength{\tabcolsep}{0pt}
\begin{tabular}{|p{1cm}|c|c|c|c|c|c|c|c|c|c|}

\hline
& Train on & \multicolumn{8}{c|}{DA} & Train on\\
& source& \multicolumn{2}{c|}{Unsupervised} & \multicolumn{2}{c|}{Labeled from source only} & \multicolumn{4}{c|}{ Labeled from source and target} &  target  \\
& No adapt. & SGF~\cite{gopalan11}$^*$ & GFK~\cite{gong12}$^*$ & GFK~\cite{gong12}$^*$ & OT-lab~\cite{Cou14}$^*$ &GFK~\cite{gong12}$^\dagger$&MMDT~\cite{Hof13}$^\dagger$& KEMA $K_{i}$ & KEMA $K_{\chi_2}$ &   No. adapt\\\hline
$l_S$ & 0 & 0 & 0 & 20 & 20 & 20 & 20 & 20 &   20 &  0 \\
$l_T$ & 0 & 0 & 0 & 0 & 0 & 3 & 3 & 3 & 3 &  8 \\
\hline
\texttt{C} $ \to $ \texttt{A} & $21.4 \pm 3.7$ & 
$36.8 \pm 0.5$ & $36.9 \pm 0.4$ & $40.4 \pm 0.7$ & $43.5 \pm 2.1$ &
$44.7 \pm 0.8$& 
$49.4 \pm 0.8$& 
$47.1 \pm 3.0$&  
$47.9 \pm 3.2$&  
$35.4 \pm 2.4$\\ 

\texttt{C} $ \to $ \texttt{D} & $12.3 \pm 2.8$ & 
$32.6 \pm 0.7$ & $35.2 \pm 1.0$ & $41.1 \pm 1.3$ & $41.8 \pm 2.8$ &
$57.7 \pm 1.1$& 
$56.5 \pm 0.9$& 
$61.5 \pm 2.8$&
$63.4 \pm 3.4$&   
$65.1 \pm 1.9$\\ 

\texttt{A} $ \to $ \texttt{C} & $19.9 \pm 1.9$& 
$35.3 \pm 0.5$ & $35.6 \pm 0.4$ & $37.9 \pm 0.4$ & $35.2 \pm 0.8$ &
$36.0 \pm 0.5$& 
$36.4 \pm 0.8$& 
$29.5 \pm 3.0$& 
$30.4 \pm 3.3$&   
$28.4 \pm 1.6$\\ 

\texttt{A} $ \to $ \texttt{W} & $17.5 \pm 3.7$& 
$31.0 \pm 0.7$ & $34.4 \pm 0.9$ & $35.7 \pm 0.9$ & $38.4 \pm 5.4$ 
&
$58.6 \pm 1.0$& 
$64.6 \pm 1.2$& 
$65.4 \pm 2.7$& 
$66.5 \pm 2.9$&   
$63.5 \pm 2.6$\\

\texttt{W} $ \to $ \texttt{C} & $24.2 \pm 1.4$& 
$21.7 \pm 0.4$ & $27.2 \pm 0.5$ & $29.3 \pm 0.4$ & $35.5 \pm 0.9$ 
 &
$31.1 \pm 0.6$& 
$32.2 \pm 0.8$& 
$32.9 \pm 3.3$&
$32.4 \pm 3.0$&   
$28.4 \pm 1.6$\\ 

\texttt{W} $ \to $ \texttt{A} & $27.0 \pm 1.5$& 
$27.5 \pm 0.5$ & $31.1 \pm 0.7$ & $35.5 \pm 0.7$ & $40.0 \pm 1.0$
 &
 $44.1 \pm 0.4$& 
$47.7 \pm 0.9$& 
 $44.9 \pm 4.5$& 
$45.9 \pm 3.9$&   
$35.4 \pm 2.4$\\ 

\texttt{D} $ \to $ \texttt{A}& $19.0 \pm 2.2$& 
$32.0 \pm 0.4$ & $32.5 \pm 0.5$ & $36.1 \pm 0.4$ & $34.9 \pm 1.3$
&
$45.7 \pm 0.6$& 
$46.9 \pm 1.0$& 
 $44.2 \pm 3.1$& 
$45.2 \pm 3.4$&   
$35.4 \pm 2.4$\\ 

\texttt{D} $ \to $ \texttt{W} & $37.4 \pm 3.0$& 
$66.0 \pm 0.5$ & $74.9 \pm 0.6$ & $79.1 \pm 0.7$ & $ {84.2 \pm 1.0}$ 
 &
 $76.5 \pm 0.5$& 
$ 74.1 \pm 0.8$& 
 $64.1 \pm 2.9$&
$66.7 \pm 3.1$&  
$63.5 \pm 2.6$\\
\hline
{\bf Mean} & 22.34 & 35.36 & 38.48 & 41.89 & 44.19 & 49.30 & 50.98 & 48.70 & 49.80 &  44.39\\
\hline

\end{tabular}
\end{center}
\end{table*}

\subsection{Recognition of facial expressions in multi-subject databases}


This experiment deals with the task of recognizing facial expressions. We used the dataset in~\cite{Son07}, where 185 photos of three subjects depicting three facial expressions (happy, neutral and shocked) are available. Each image is $217 \times 308$ pixels and we take each pixel as one dimension for classification: the problem is $200`508$ dimensional. 
Each pair \{subject,expression\} has around $20$ repetitions. 

\begin{figure}[h!]
\begin{center}
\setlength{\tabcolsep}{-4pt}
\begin{tabular}{cccc}
(a) Error rates & (b) Subjects & (c) Labels & (d) Predictions \\
\includegraphics[width=4cm]{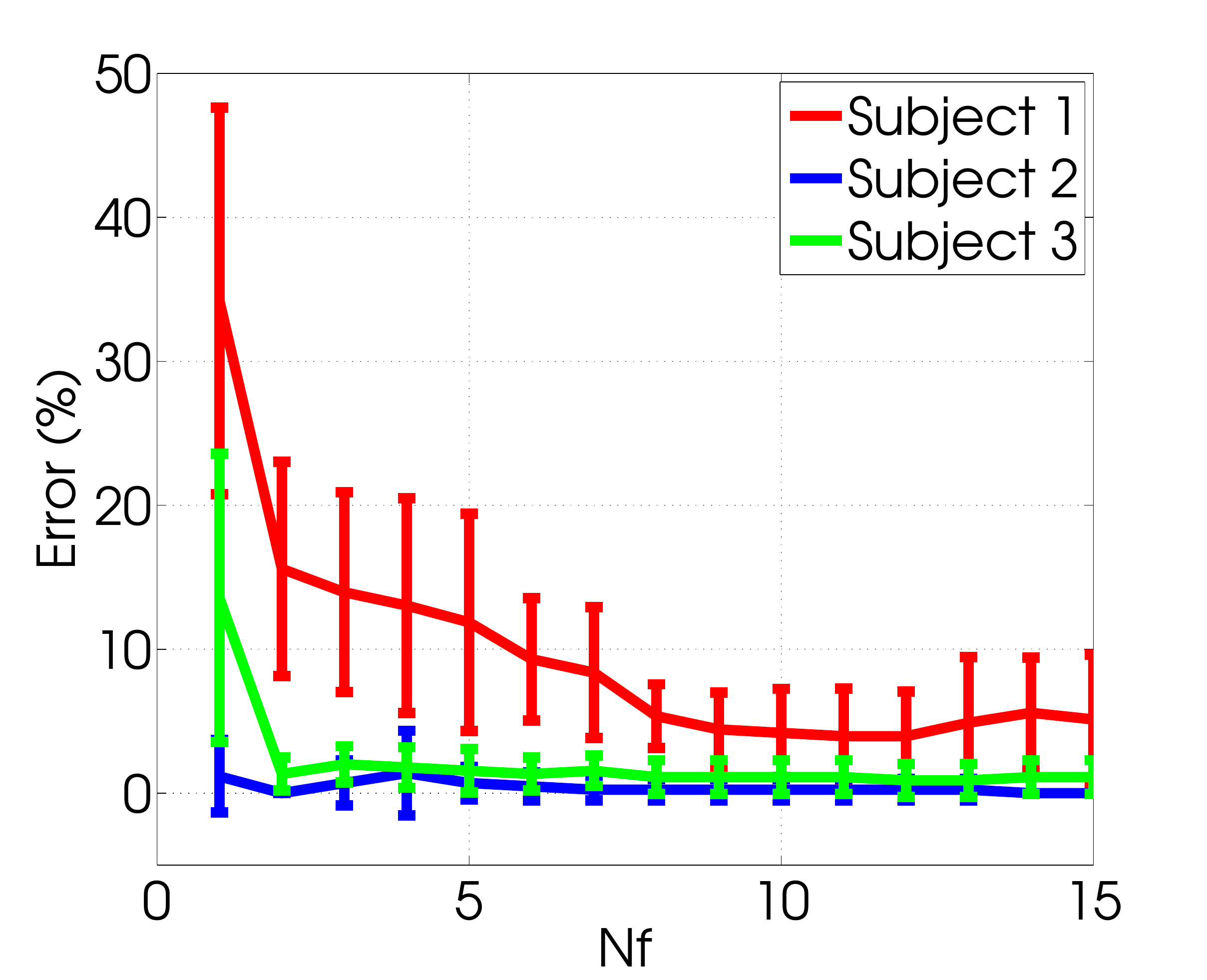}&
\includegraphics[width=4cm]{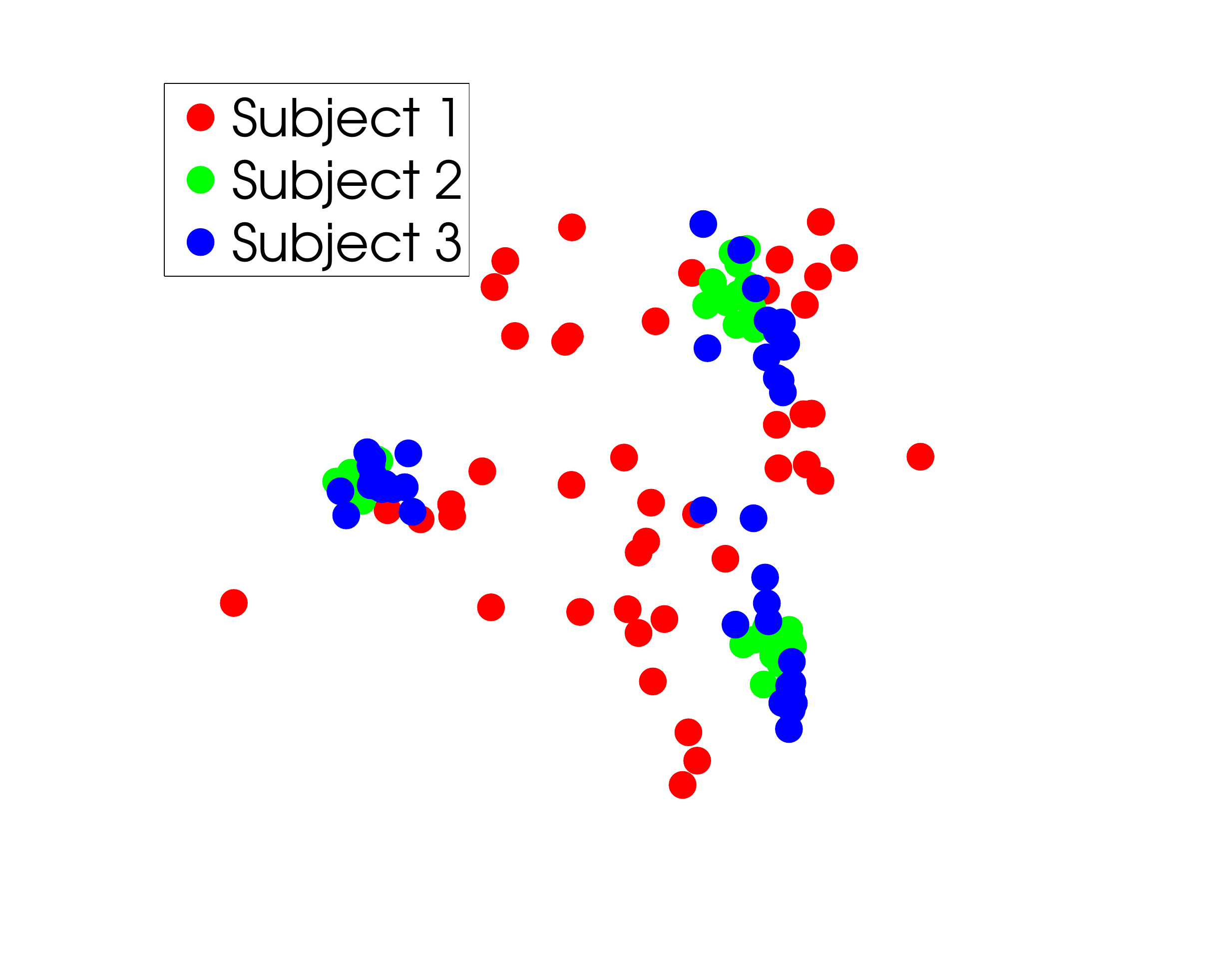} &
\includegraphics[width=4cm]{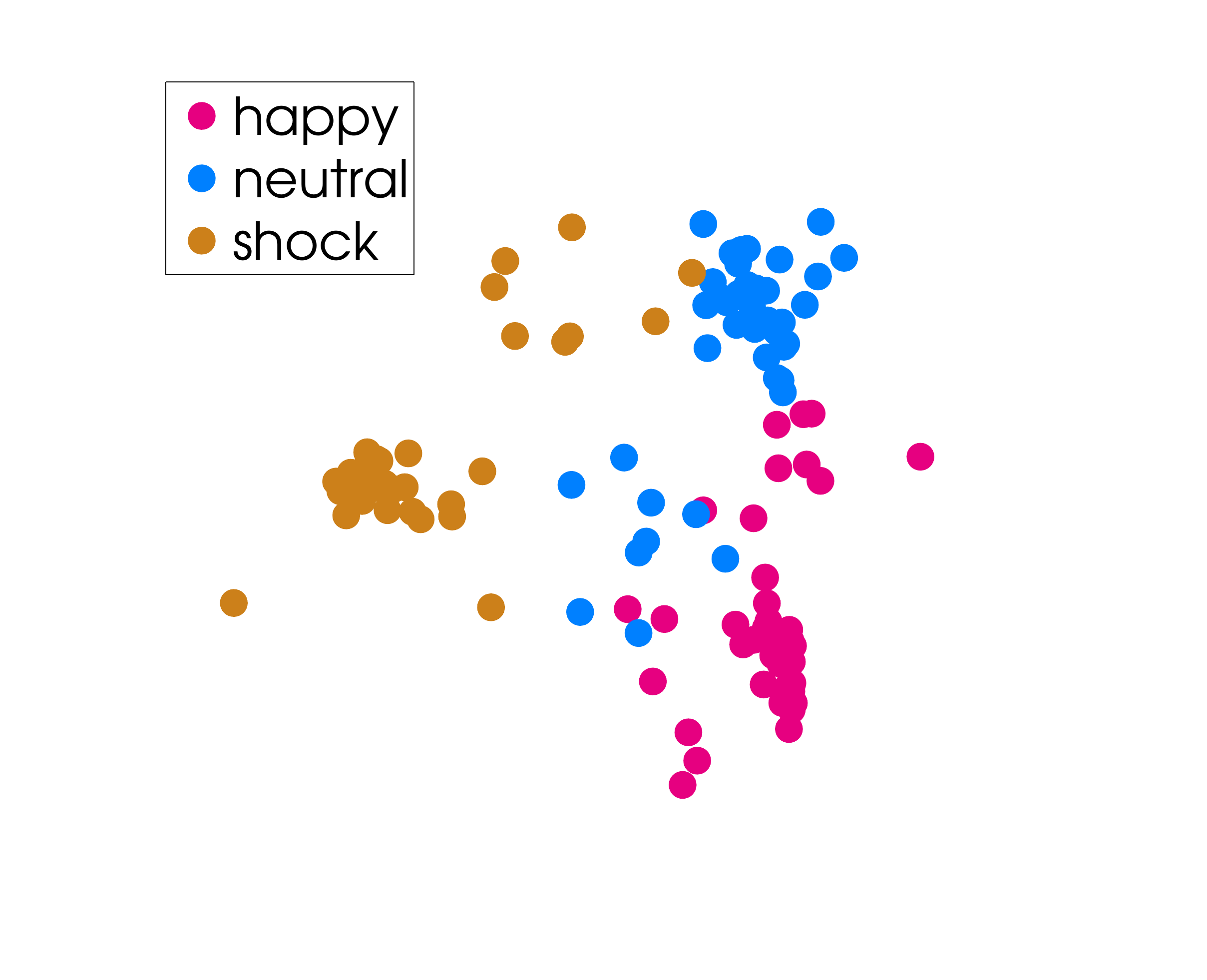}&
\includegraphics[width=4cm]{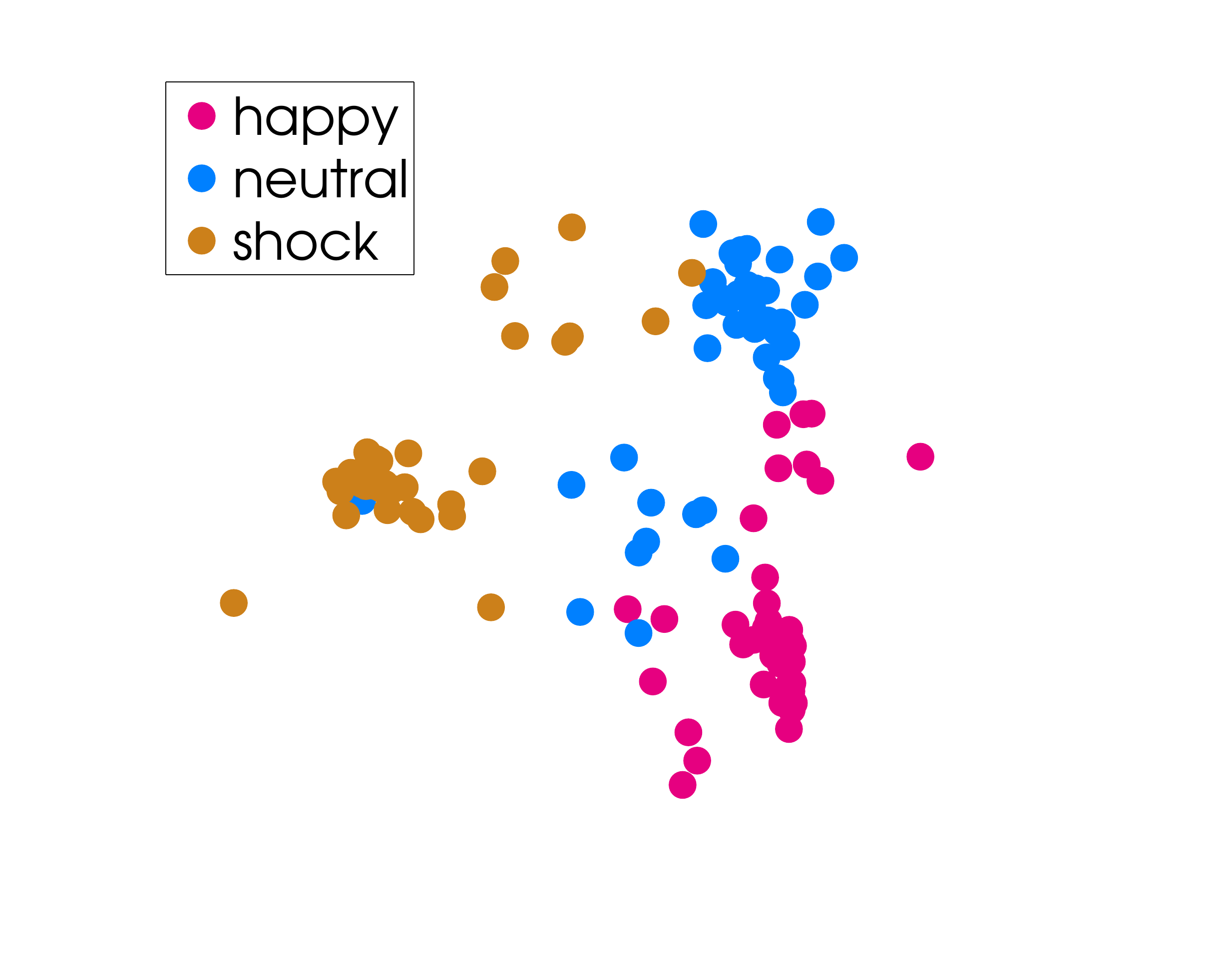}
\end{tabular}
\vspace{-0.25cm}
\caption{Results of the classification of facial expressions. 
} \label{fig:resfaces}
\end{center}
\end{figure}

Different subjects represent the domains and we align them with respect to the three expression classes. We used only three labeled examples {\em per} class and subject. 
Results are given in Fig.~\ref{fig:resfaces}(a): since it works directly in the dual, KEMA can effectively cast the three-domains problem into a single ten-dimensional latent space, where all domains are classified with less than 5\% error. This shows an additional advantage of KEMA with respect to SSMA in high dimensional spaces: SSMA would have required to solve a $601`524$-dimensional eigenproblem, while KEMA solves only a $55$-dimensional problem.  
Figures~\ref{fig:resfaces}(b)-(d) present different visualizations of the two first dimensions of the latent space: subject \#1 seems to be the most difficult to align with the two others, difficulty that is also reflected in the higher classification errors. 
Actually, subject \#1 shows  little variations in his facial traits from one expression to the other compared to the other subjects (see Fig.~3 in~\cite{Son07}).

\section{Conclusions}

We introduced a kernel method for semi-supervised manifold alignment. We want to stress that this particular kernelization goes beyond the standard academic exercise as the method addresses many problems in the literature of domain adaptation and manifold learning. The so-called KEMA can actually align an arbitrary number of domains of different dimensionality without needing corresponding pairs, just few labeled examples in all domains. We also showed that KEMA generalizes SSMA when using a linear kernel, which allows us to deal with high-dimensional data efficiently in the dual form. Working in the dual can be computationally costly because of the construction of the graph Laplacians and the size of the involved kernel matrices. Regarding the Laplacians, they can be computed just once and off-line, while regarding the size of the kernels, we introduced a reduced-ranked version that allows to work with a fraction of the samples while maintaining the accuracy of the representation. Advantageously, KEMA can align manifolds of very different structures and dimensionality, performing a sort of manifold unfolding along with the alignment. Importantly, the inversion of the KEMA projections has a closed-form solution without the need of pre-imaging. This is an important feature that allows synthesis applications, but more remarkably allows to study and characterize the distortion of the manifolds in physically meaningful units.  To authors' knowledge this is the first method in addressing all these important issues at once. All these features were illustrated through toy examples and real problems in computer vision and machine learning.


\subsubsection*{References}
\def\refname{~}\vspace{-1cm}
\bibliographystyle{unsrt}
\bibliography{kadapt}

\end{document}